\documentclass[preprint,12pt]{elsarticle}
\usepackage{graphicx} 

\usepackage{algorithm}
\usepackage{algorithmic}
\usepackage{amsmath}
\usepackage{amssymb}
\usepackage{mdwtab}
\usepackage{eqparbox}
\usepackage{makecell}
\usepackage{tabularray}
\usepackage[hidelinks]{hyperref}
\usepackage{comment}
\usepackage{lscape}
\usepackage{multicol}
\usepackage{array}

\hypersetup{
    colorlinks=true,
    linkcolor=blue,
    filecolor=blue,      
    urlcolor=blue,
    citecolor=blue,
}

\usepackage[pdftex]{color}
\usepackage[font=footnotesize,labelfont=bf]{caption}
\usepackage[font=footnotesize,labelfont=bf]{subcaption}
\affiliation[label1]{organization={Concordia Institute for Information Systems Engineering, Concordia University},
            city={Montreal}, 
            country={Canada}}

\affiliation[label3]{organization={Department of Mechanical, Industrial and Aerospace Engineering, Concordia University},
            city={Montreal}, 
            country={Canada}}

\affiliation[label4]{organization={Department of Medicine, Laval University},
            city={Quebec}, 
            country={Canada}}

\affiliation[label7]{organization={Department of Electrical, Computer, and Software Engineering, Ontario Tech University}, city={Oshawa}, country={Canada}}

\affiliation[label5]{organization={Department of Computer Science, Khalifa University},
            city={Abu Dhabi}, 
            country={UAE}}

\affiliation[label6]{organization={Artificial Intelligence \& Cyber Systems Research Center, Department of CSM, Lebanese American University}, city={Beirut}, country={Lebanon}}

\affiliation[label8]{organization={Department of Electrical Engineering and Computer Science, Khalifa University}, city={Abu Dhabi}, country={UAE}}

\author[label1,label5]{Hanae Elmekki} 
\ead{hanae.elmekki@mail.concordia.ca}

\author[label3]{Amanda Spilkin} 
\ead{amanda.spilkin@mail.concordia.ca}

\author[label3]{Ehsan Zakeri} 
\ead{ehsan.zakeri@concordia.ca}

\author[label4]{Antonela Mariel Zanuttini}
\ead{antonela-mariel.zanuttini.1@ulaval.ca}

 \author[label1]{Ahmed Alagha}
 \ead{ahmed.alagha@mail.concordia.ca}

\author[label7]{Hani Sami}
\ead{hani.sami@ontariotechu.ca}

\author[label5,label1]{Jamal Bentahar \corref{cor1}}
\cortext[cor1]{Corresponding Author's Email:  jamal.bentahar@ku.ac.ae}
\ead{jamal.bentahar@ku.ac.ae}

\author[label3]{Lyes Kadem}
\ead{lyes.kadem@concordia.ca}

\author[label3]{Wen-Fang Xie}
\ead{wenfang.xie@concordia.ca}

\author[label4]{Philippe Pibarot}
\ead{philippe.pibarot@med.ulaval.ca}

\author[label5]{Rabeb Mizouni}
\ead{rabeb.mizouni@ku.ac.ae}

\author[label5]{Hadi Otrok}
\ead{hadi.otrok@ku.ac.ae}

\author[label5,label6]{Azzam Mourad}
\ead{azzam.mourad@lau.edu.lb}

\author[label8]{Sami Muhaidat} 
\ead{sami.muhaidat@ku.ac.ae}
            
\date{July 2025}

\begin{document}

\begin{frontmatter}

\title{End-to-End Framework Integrating Generative AI and Deep Reinforcement Learning for Autonomous Ultrasound Scanning}

\begin{abstract}

Cardiac ultrasound (US) is among the most widely used diagnostic tools in cardiology for assessing heart health, but its effectiveness is limited by operator dependence, time constraints, and human error. The shortage of trained professionals, especially in remote areas, further restricts access. These issues underscore the need for automated solutions that can ensure consistent, and accessible cardiac imaging regardless of operator skill or location. Recent progress in artificial intelligence (AI), especially in deep reinforcement learning (DRL), has gained attention for enabling autonomous decision-making. However, existing DRL-based approaches to cardiac US scanning lack reproducibility, rely on proprietary data, and use simplified models. Motivated by these gaps, we present the first end-to-end framework that integrates generative AI and DRL to enable autonomous and reproducible cardiac US scanning. The framework comprises two components: \textbf{(i)} a conditional generative simulator combining Generative Adversarial Networks (GANs) with Variational Autoencoders (VAEs), that models the cardiac US environment producing realistic action-conditioned images; and \textbf{(ii)} a DRL module that leverages this simulator to learn autonomous, accurate scanning policies. The proposed framework delivers AI-driven guidance through expert-validated models that classify image type and assess quality, supports conditional generation of realistic US images, and establishes a reproducible foundation extendable to other organs. To ensure reproducibility, a publicly available dataset of real cardiac US scans is released. The solution is validated through several experiments. The VAE-GAN is benchmarked against existing GAN variants, with performance assessed using qualitative and quantitative approaches, while the DRL-based scanning system is evaluated under varying configurations to demonstrate effectiveness.

\end{abstract}


\begin{keyword}
Ultrasound Imaging \sep Cardiac Scanning \sep Generative Adversarial Network
 \sep Reinforcement Learning \sep Image Generation \sep Robotics.
\end{keyword}
\end{frontmatter}


\section{Introduction}
\label{Sec:Introduction}

Echocardiography, also known as cardiac ultrasound (US), is one of the primary techniques for assessing cardiac health due to its several advantages: it is non-invasive, provides real-time imaging, and is widely used for evaluating both the structure and function of the heart \cite{wang2021deep}. In a standard transthoracic echocardiography (TTE) examination, the individual performing the scan, whether a sonographer, medical imaging specialist, or cardiologist can vary depending on the healthcare system and country \cite{yu2017challenges}. The examination often begins with standard parasternal long-axis views, followed by parasternal short-axis, apical views, and Doppler imaging. The scanning protocol may vary depending on the suspected cardiac pathology, and certain imaging windows may require more time and precision than others.

TTE procedure is inherently complex, as it involves scanning multiple cardiac windows and views, requiring a high degree of precision and the expertise of skilled, well-trained operators to ensure proper alignment during image acquisition, which is essential for accurate quantitative measurements and effective treatment monitoring \cite{yu2017challenges}. US scanning in general is also considered time-consuming, often leading to musculoskeletal disorders and operator fatigue due to prolonged physical strain \cite{zangiabadi2024musculoskeletal}. In addition to these challenges, rural and remote areas frequently face a shortage of trained US professionals, which limits access to timely and accurate diagnoses \cite{adams2021access}. These limitations have prompted ongoing research efforts aimed at improving the accessibility and efficiency of echocardiography. In this context, artificial intelligence (AI), particularly deep learning (DL) techniques, has emerged as a promising solution. DL has seen rapid advancement and adoption in the medical domain, especially in cardiovascular applications, where it supports clinical decision-making and enhances healthcare service delivery \cite{wang2021deep, petmezas2024recent, chen2020deep}.

In the context of US scanning, DL is used not only to address traditional computer vision tasks such as classification and segmentation, but also to automate the scanning process itself, an advancement enabled in part by progress in the field of robotics \cite{jiang2023robotic}. Indeed, the development of autonomous robotic systems for US scanning has become an active area of research, as it addresses many of the challenges associated with manual US procedures. Numerous studies in the literature have explored this application across various human organs, including the liver, thyroid, and others \cite{bi2024autonomous, su2024fully,li2021autonomous, ao2025saferplan}. The heart has also emerged as a target organ for autonomous US scanning \cite{amadou2024goal, li2023rl, lin2023deep, shen2023towards}. Most of these approaches rely on deep reinforcement learning (DRL), a subset of DL that enables an agent to learn how to reach a target through interaction with its environment, guided by a reward-penalty mechanism to optimize its decision-making policy. DRL has demonstrated strong performance in this domain and has emerged as a promising alternative to other strategies such as imitation learning \cite{dall2024imitation} and visual servoing \cite{zhou2024fully}. However, one of the main challenges in applying DRL to US scanning is the need for a realistic simulation environment to train the DRL models. 

One major issue is the scarcity of data required to construct realistic simulation environments. Unlike many other fields, medical data, including US images, is often limited due to privacy concerns and ethical constraints. This limited availability significantly hinders the development of accurate and diverse simulation frameworks. Moreover, even when real data is available, generating realistic US images is a complex task. Many simulation approaches rely on overly simplified representations of the scanned region, such as binary or grid-based models \cite{bi2022vesnet,hase2020ultrasound}. These approaches may be insufficient to capture the intricate tissue characteristics necessary for high-quality training. Although some methods attempt to address this by synthesizing panoramic views from labeled image sequences \cite{su2024fully} or by leveraging data from other imaging modalities such as computed-tomography scans, these approaches often do not fully replicate the complex tissue-probe interactions and visual artifacts present in real US imaging \cite{lin2023deep, amadou2024goal}. 

Another consideration in DRL  for US automation is the definition of the state space, action space, and reward function. In existing work focused on automating US cardiac scanning using DRL, various strategies have been proposed for state representation. For instance, in transesophageal echocardiography (TEE), some methods use the pose of the US transducer as the state input \cite{amadou2024goal}, while others use the US image itself \cite{li2023rl}. In TTE, the state has been defined in multiple ways, such as the relative positions of reference points from the current and target US views \cite{lin2023deep}, 2D US images extracted from a 3D volume \cite{shen2023towards}, or a combination of probe position and heart detection confidence \cite{shida2024robotic}. However, none of these approaches rely solely on the visual feedback from the current US image during scanning. This is notable because spatial coordinates may vary significantly between patients, and in many clinical scenarios, a predefined target image is not available.

Regarding the action space, different works adopt different approaches. For TTE, \cite{shida2024robotic} defines a set of discrete actions (up, down, left, right, diagonals, and stop), while \cite{shen2023towards} models eight actions that adjust the orientation and position of the 2D plane within a 3D volume. In contrast, \cite{lin2023deep} controls the robot by increasing or decreasing the values of its four joints. However, none of these works explicitly model the full set of 12 possible actions corresponding to the six degrees of freedom (DoF) of the US probe—comprising translations and rotations along the three principal axes (X, Y, Z), each with positive and negative directions.

Reward function design also varies significantly across studies. For instance, \cite{amadou2024goal} provides a reward based on the proximity of the transducer to the goal state. In \cite{lin2023deep}, the agent receives rewards for actions that minimize a loss function and is penalized otherwise. Similarly, \cite{shen2023towards} quantifies differences between the target and current planes using visual and geometric parameters, while \cite{li2023rl} assigns a discrete reward for reaching the goal, a penalty for exceeding movement limits, and otherwise uses a weighted sum of rewards based on position, orientation, and compliance. However, none of the existing works provide a reward signal based solely on continuous visual feedback from the currently observed US image, nor do they evaluate the quality of the scanned image during the scanning process itself.

To summarize, the key challenges in current applications of DRL for autonomous cardiac US scanning are:

\begin{itemize}
\item Limited availability of medical US data due to ethical and privacy constraints.
\item Oversimplified simulation models lack the anatomical detail required for effective model training and are not suitable for complex organs such as the heart.
\item Difficulty in reproducing and generalizing existing simulation environments to different organs.
\item Lack of DRL approaches that rely solely on continuous visual feedback and use an enhanced action space covering the full 6 DOF of the US probe for the cardiac scanning.
\item Existing reward functions in DRL applications for cardiac US scanning do not incorporate continuous evaluation of image quality.
\end{itemize}

To address these challenges, this work proposes a DRL-based robotic system capable of autonomously and accurately performing cardiac US scans. The system consists of two main components: \textbf{(1)} a GAN-based image generation module for simulating the training environment, and \textbf{(2)} a DRL-based US scanning module. To create a reproducible simulation environment for DRL applications, generative AI is employed, particularly Generative Adversarial Networks (GANs), as a solution for generating realistic cardiac US images conditioned on US probe and robotic control parameters. The specific approach used for generating realistic US images is the Conditional Generative Adversarial Network (cGAN). cGANs have been widely applied in medical imaging tasks, including image-to-image translation \cite{kodipalli2022segmentation, gu2021echocardiogram, long2024spatial}, image resolution enhancement \cite{liu2025reconstruction}, and 3D reconstruction \cite{sun2022hierarchical}. In this work, we leverage a cGAN model to generate realistic cardiac US images based on robotic and probe-related conditions. To further improve the uniqueness and quality of the generated images, we integrate a Variational Autoencoder (VAE) into the GAN framework, enabling the model to better capture the variability and structural complexity found in real cardiac US data. 

Building on this simulated environment, we propose a DRL-based framework to autonomously guide a robotic system for cardiac US scanning. Unlike previous approaches, our framework is primarily designed for cardiac imaging; however, it is extensible, generalizable, and reproducible across other organs with appropriate data and parameter tuning. A key contribution of this work is the integration of a DL-based image quality assessment module into the DRL reward function. This enables the robotic agent not only to reach anatomically correct views but also to prioritize diagnostically significant ones. Due to the scarcity of human cardiac US data required for developing DRL-based solutions, we also created a publicly available dataset consisting of US images acquired from  a Phantom simulating the heart along with their corresponding spatial and robotic parameters, which can be reused by the research community. Overall, the main contributions of this work are summarized as follows:

\begin{itemize}
\item Development of a realistic simulation framework for US image generation using generative AI, conditioned on spatial and robotic parameters.
\item Design and implementation of a DRL-based system for autonomous cardiac US scanning using a state space based solely on visual feedback, with an enhanced action space and a reward function to evaluate the anatomical accuracy and image quality of the scanned US.
\item Benchmarking of multiple generative AI models, evaluating their performance in generating high-quality and diverse US images.
\item Creation of a publicly available dataset \textbf{RACINES} (Robotic Acquisition for Cardiac Intelligent Navigation Echography Systems) for cardiac US scanning to support reproducibility and further research in autonomous US systems.
\end{itemize}

The remainder of this paper is structured as follows: Section \ref{Sec: RelatedWork} provides an overview of current work in medical image generation and autonomous US scanning. Section \ref{Sec: Proposed Framework} introduces the proposed framework. This is followed by a detailed description of the GAN-based simulation environment in Section \ref{subsec: GAN_environment_simulation}, and the DRL-based autonomous US scanning in Section \ref{Sec: drl_based_autonomous_ultrasound_scanning}. Section \ref{Sec: Experiments} presents the experimental results, and finally, Section \ref{Sec: concolusion} concludes the paper.

\section{Related Work}
\label{Sec: RelatedWork}

The objective of this section is to explore and review recent advancements in the fields of medical image generation and autonomous US scanning. We are particularly interested in medical image generation using GANs because it is impractical to collect data for every possible point on a phantom model as there are infinitely many positions to consider. GANs offer a solution by generating realistic medical images for those missing or unobserved points, thereby enriching the dataset used to simulate the environment for the proposed DRL approach. In this section, we will review the current literature on medical image generation, focusing on both traditional GANs and cGANs, the latter of which generate images based on specific input conditions. Following this, we will examine the state of the art in autonomous US scanning, including applications in cardiac imaging as well as other organs.

\subsection{Medical Image Generation}
\label{subsec: generation datasets}

The application of AI in the medical field is gaining increasing attention due to its potential to assist in clinical decision-making. However, one of the primary challenges in this domain is the scarcity of medical imaging data, which is difficult to collect due to ethical concerns and regulatory constraints. To address this limitation, data augmentation techniques have traditionally been employed to artificially expand the size and variability of training datasets. However, conventional augmentation methods 
only alter the appearance of existing images and are unable to create entirely new, realistic samples. This limitation has led to growing interest in more advanced generative approaches, particularly GANs.

GANs have emerged as a promising solution for various applications in medical imaging, especially for tasks such as image classification \cite{kodipalli2022segmentation} and segmentation \cite{alruily2023breast,shi2025semantic, DBLP:journals/mia/YiWB19}. While GANs have gained significant attention for their ability to generate realistic images, their capacity to perform conditional image generation where outputs are guided by specific inputs such as images, numerical features, or semantic labels has further amplified interest in this field. The work by Mirza et al. \cite{mirza2014conditional} represents a pioneering contribution to the field, introducing cGANs, which incorporate discrete conditioning variables to guide the image generation process.

For example, in \cite{wang2024us2mask}, a cGAN is employed to generate multi-class rib segmentation masks from cardiac US images. The generator learns to produce accurate masks based on the input US image, which is a significant contribution given that manual annotation of segmentation datasets is time-consuming and labor-intensive. Similarly, in \cite{jwaid2024elevating}, the authors perform the reverse task by synthesizing cardiac US images conditioned on binary semantic masks. Likewise, in \cite{datta2024conditional}, GANs are used to generate segmented US images based on a combination of features, including raw image information, segmentation masks, and noise vectors. This supports breast lesion diagnosis by facilitating image segmentation. 
In another example, the authors of \cite{shi2025semantic} proposed a GAN model in which the generator is implemented using an encoder–decoder architecture with residual connections and channel attention mechanisms. 
The model is conditioned on both segmentation masks and edge semantic sketches, providing the generator with richer contextual information compared to models conditioned solely on segmentation masks. This approach aligns with the method proposed in \cite{qiao2022pseudo}, where a sketch image is used as a condition to synthesize high-resolution fetal cardiac US images. 

cGANs are frequently used in image-to-image translation tasks, where the input (condition) and output are both images. For instance, \cite{gu2021echocardiogram} uses a cGAN to translate an apical four-chamber cardiac image into an apical two-chamber view. The generator employs a U-Net architecture, and a segmentation network is integrated to enhance robustness and guide the transformation. Similarly, \cite{kodipalli2022segmentation} translates raw pelvic images into segmented images highlighting ovarian tumors, which are then passed through a classifier to determine malignancy. 

Beyond 2D imaging, GANs have also been applied in 3D medical imaging. However, due to the high computational cost of generating full 3D volumes, researchers have adopted alternative strategies. For instance, \cite{sun2022hierarchical} introduces a cGAN to generate low-resolution 2D slices and 3D sub-volumes of CT images based on discrete text prompts. This model combines a GAN with an autoencoder, enabling the generator to function both as an image creator and a reconstructor. Similarly, in \cite{jung2021conditional}, 2D brain slices are synthesized and assembled into 3D brain volumes based discrete conditions representing Alzheimer’s disease stages.

cGANs are also used to enhance image resolution. In \cite{liu2025reconstruction}, a cGAN with an attention-enhanced U-Net generator and a convolutional neural network (CNN)-based discriminator is used to generate high-resolution US images from low-resolution inputs. Another related approach, EchoGAN, is developed for image outpainting, specifically to expand the field of view in TTE. This addresses the issue where increasing image resolution typically reduces the visible field. EchoGAN uses a U-Net-based generator and a CNN-based discriminator, conditioned on a binary mask.

This literature review highlights that the use of cGANs with continuous conditions is relatively rare in the field of medical imaging. Continuous Conditional GANs (ccGANs) were first introduced in \cite{ding2021ccgan}, where the conditioning variables are continuous scalar values, such as angles or age, rather than categorical labels. The label is embedded via a linear layer and projected into the discriminator’s intermediate feature space. However, this approach does not fully align with the context of our  work, where labels are high-dimensional vectors with varying ranges (e.g., 12 distinct continuous values). Moreover, the richness of content in US images and the complexity of their continuous acquisition conditions pose significant challenges to the effectiveness of traditional GAN architectures, highlighting the need for more advanced generative models capable of accurately simulating the US environment. In particular, for training DRL-based solutions, simulation environments must be as realistic as possible to closely replicate real-world conditions and ensure high DRL performance. In the literature, many simulation approaches rely on simplified representations of the scanned region, such as binary or grid-based models \cite{bi2022vesnet,hase2020ultrasound}. These approaches may be insufficient to capture the intricate tissue characteristics necessary for high-quality training. Although some methods attempt to address this by synthesizing panoramic views from labeled image sequences \cite{su2024fully} or by leveraging data from other imaging modalities such as computed-tomography scans, these approaches often do not fully replicate the complex tissue-probe interactions and visual artifacts present in real US imaging \cite{lin2023deep, amadou2024goal}. Using conditional generative AI will presents a significant research opportunity, as such simulations could facilitate training intelligent agents in a controlled and reproducible manner, potentially accelerating advancements in autonomous US scanning systems.

\subsection{Autonomous US Scanning}
\label{subsec: autonomous scanning}

Autonomous US scanning has recently emerged as one of the most prominent research areas in medical robotics. This section reviews current approaches to autonomous US scanning, emphasizing recent advancements and identifying key limitations, especially in cardiac imaging, through an analysis of how DRL methods have been designed and implemented.

An early approach to robotic US scanning involves tele-echography systems \cite{si2024design, wang2021application}, in which expert sonographers remotely control robotic arms to perform US scans. However, these systems lack autonomy, which limits their scalability and effectiveness, particularly in scenarios where expert supervision is unavailable. Other approaches leverage visual servoing \cite{zhou2024fully, welleweerd2020automated, zakeri2025robust} or marker-based techniques \cite{pan2017comparison,rosen2011surgical}, using visual feedback from external cameras to guide robotic movement. While effective in controlled environments, these methods often lack adaptability and intelligence, making them unsuitable for real-world clinical deployment where anatomical variability is significant.

From the perspective of AI, recent research has introduced imitation learning (IL) as a way to teach robots how to perform US scanning. For instance, Dall et al. \cite{dall2024imitation} applied IL to train a robotic system to diagnose Deep Vein Thrombosis (DVT) using expert demonstrations. Similarly, Wang et al. \cite{wang2024robotic} developed a semi-autonomous system in which a human operator remotely guides the probe near the acoustic window, after which the robot autonomously fine-tunes the probe’s position to achieve the desired cardiac view. Other studies, such as \cite{si2025deep}, apply IL to train robots to scan the carotid artery. Beyond IL, Wu et al. \cite{wu2025point} proposed a point-cloud-based approach for kidney scanning. Using a depth camera to collect 3D data of a patient's back surface, they identify anatomical landmarks to localize the kidney. Another approach includes path planning using time-invariant dynamical systems \cite{liuchen2024dynamical} and the use of virtual fixtures combined with segmentation techniques to guide scanning for DVT \cite{huang2024robot}. Despite their contributions, these approaches share common limitations: they often lack continuous feedback mechanisms, adaptability to anatomical variations, and robust exploration capabilities. These shortcomings have motivated the exploration of DRL as a powerful alternative. DRL enables systems to learn optimal scanning strategies through direct interaction with the environment, thereby addressing many of the constraints faced by manual or rule-based methods.


Several studies have applied DRL to autonomous US scanning across various organs, including the kidney, spine, and heart \cite{elmekki2025comprehensive}. These works utilize different DRL algorithms, such as Deep Q-Networks (DQN), and exhibit considerable variation in their state representations and reward formulations. In the context of cardiac imaging, for example, \cite{amadou2024goal} developed a DRL-based system for TEE using an actor-critic architecture. In this setup, the agent observes US images and performs actions such as translating the probe along the esophagus and rotating the transducer. Similarly addressing TEE, \cite{li2023rl} employs a DQN-based approach that considers the US image as the state. For TTE, \cite{lin2023deep} adopts a DQN, where the state is defined as the relative positions of three reference points in the current US view compared to the target view. The reward function encourages actions that minimize a loss function and penalizes those that increase it. In another approach, \cite{shen2023towards} applies a Dueling Double DQN using a single US image as the state, with rewards computed based on the similarity between the current and target scan planes.

Beyond cardiac imaging, recent works have also explored DRL-based scanning for other organs. For kidney imaging, Xu et al. \cite{xu2025efficient} propose a composite state representation that includes US image sequences, probe camera inputs, force feedback, and action history. The reward function integrates multiple factors, including probe positioning accuracy, image quality, and applied force. For vascular imaging, Medany et al. \cite{medany2025model} define the state as a combination of raw US images and the spatial coordinates of the probe, with an action space that includes control over the probe’s vibration frequency and amplitude. Additionally, Li et al. \cite{li2025image} employ inverse reinforcement learning to acquire expert-level scanning strategies for blood vessel imaging, using similarity metrics between the current and target images as the reward signal to assess scan quality. For more detailed insights into DRL applications across a broader range of organs and scanning scenarios, this comprehensive review \cite{elmekki2025comprehensive} can serves as a reference.

While these studies demonstrate promising results, they often lack reproducibility and generalizability. Few works provide detailed descriptions of how their simulation environments are constructed or how their methods can be adapted to different organs or scanning scenarios. Moreover, the approaches developed for cardiac US scanning tend to be significantly simplified, particularly in terms of action space and reward function design, falling short of capturing the complexity required for real-world applications. This gap motivates our work, which aims to develop a fully reproducible DRL-based framework for autonomous US scanning from simulation environment creation to scanning execution. Unlike prior efforts, our framework is designed to be generalizable across various anatomical targets, not limited to cardiac imaging, subject to appropriate dataset and parameter tuning. Another key contribution is the integration of a DL-based image quality assessment system into the reward function, enabling the robot to more effectively reach diagnostically significant views. We also propose an action space for the DRL agent that encompasses all relevant probe movements, along with a state space based solely on continuous visual feedback. Additionally, we present a comparative analysis of different state representations to evaluate their impact on scanning performance.

\section{Framework Overview}
\label{Sec: Proposed Framework}

\begin{figure}[H]
    \begin{center}
    \centering
\includegraphics[width=0.7\linewidth]{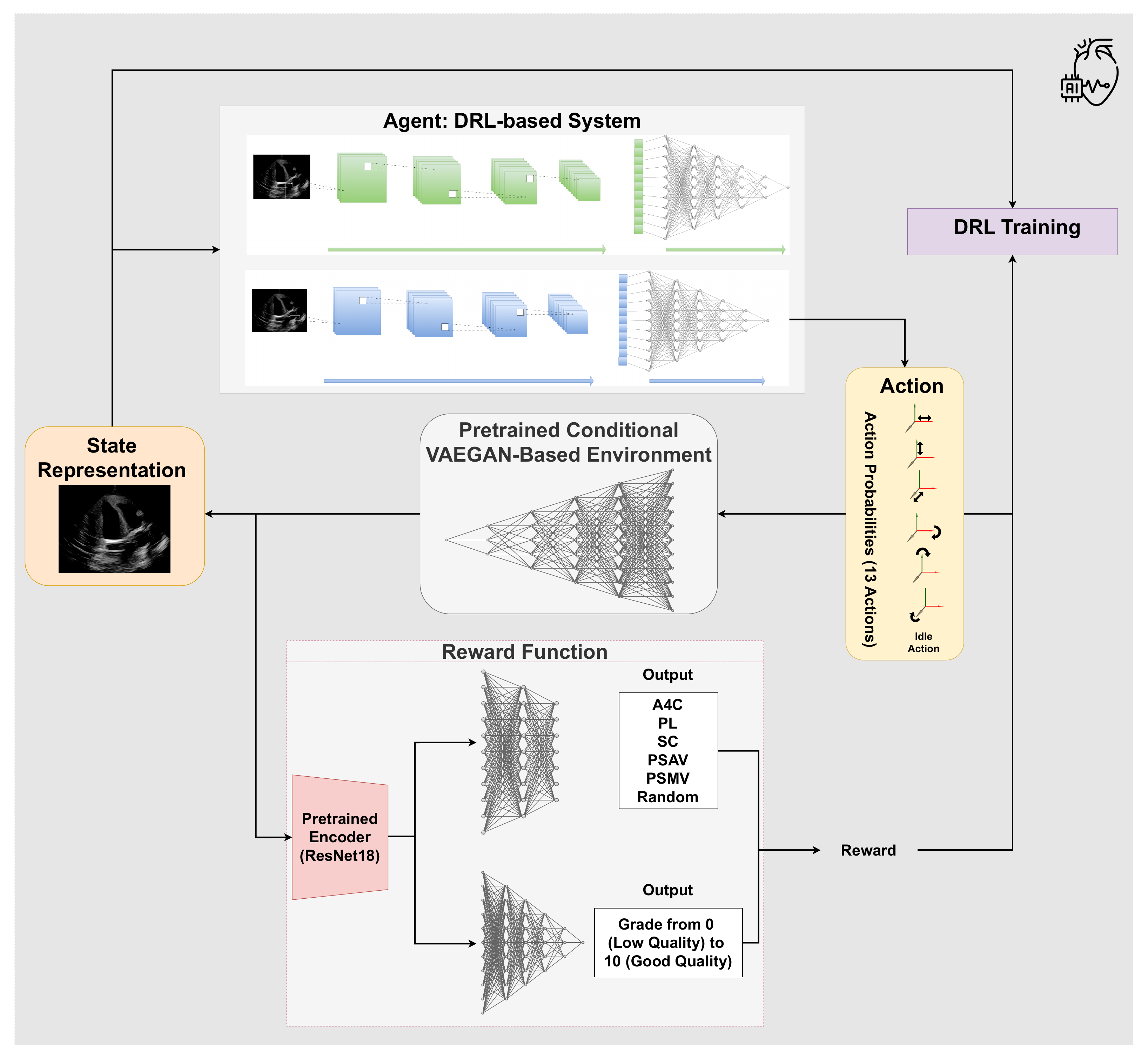}
   \caption{Overall process of the proposed autonomous US solution}
    \label{fig:framework_overview}
    \end{center}
    \vspace{-2em}
\end{figure}

The solution proposed in this study  aims to mitigate the lack of sonographers in underserved rural regions by proposing a DRL-based robotic system that can autonomously and accurately conduct cardiac US scans. An overview of the proposed solution is shown in Figure \ref{fig:framework_overview}, which consists of two main components: \textbf{(1)} a GAN-based image generation system and \textbf{(2)} a DRL-based US scanning system.

To develop a robust and accurate DRL-based US scanning system, the GAN-based generation module was designed to simulate the scanning environment, i.e. the patient’s heart region, which in this study is represented by a phantom that simulates human cardiac anatomy. This simulated environment allows the DRL system to be trained and tested until optimal performance is achieved. 

The development of a simulation environment is driven by several key factors: \textbf{1)} DRL applications require a large number of trial-and-error interactions. In the context of US scanning, a simulated environment is therefore essential to enable the safe and efficient training of DRL-based solutions, particularly for medical applications where safety and precision are critical. \textbf{2)} beyond training, the simulation environment provides a controlled setting to rigorously test and validate the performance of these DRL methods. \textbf{3)} There is a significant gap in the existing literature regarding reproducible and heart-specific US simulation environments.

The simulation is constructed by integrating a variational autoencoder (VAE) \cite{kingma2013auto} with a cGAN, conditioned on a 12-dimensional input continuous vector. The motivation for using a VAE lies in its ability to learn the latent representation of the US images, which enhances the capacity of the GAN-generator to produce realistic and diverse images that can effectively compete with the discriminator. A detailed description of this process is provided in Section \ref{subsec: Continous Conditional Ultrasound Image Generation}. In this work, the input to the VAE-GAN framework includes probe position, orientation, and robotic parameters such as force and torque. Based on these conditions, the model generates synthetic cardiac US images, which represent the state input for the DRL agent. The DRL agent is trained using the Proximal Policy Optimization (PPO) algorithm within an actor-critic framework \cite{schulman2017proximal}, which learns a policy to output a distribution over 12 possible actions. These actions correspond to translations along the x, y, and z axes and rotations around the x, y, and z axes. The choice of PPO is motivated by its stability and sample efficiency in control tasks such as US scanning. The DRL architecture is detailed and discussed in Section \ref{Sec: drl_based_autonomous_ultrasound_scanning}.

A custom reward function was designed to guide the learning process. This reward is derived from an image quality assessment model based on the ResNet-18 network \cite{he2016deep}. The grading model is implemented using transfer learning from the classification model, both of which are detailed in \cite{elmekki2025cactus}. The reward mechanism serves a dual purpose: to identify the specific cardiac view captured by the agent and to assess the image quality. Thus, the agent is encouraged not only to reach the correct anatomical view but also to generate high-quality US images. The following sections provide a detailed explanation of the design and implementation of each component of the proposed solution.

\section{GAN-Based DRL Simulation Environment}
\label{subsec: GAN_environment_simulation}

Autonomous US scanning is a complex task that requires extensive training and validation prior to deployment on human subjects. Therefore, the use of simulation environments is essential for developing robust AI models and conducting rigorous evaluation. In this work, we use the CAE Blue Phantom\footnote{\url{https://elevatehealth.net/product/cardiac-ultrasound-training-block-transparent/}} which is a static cardiac simulator to collect a comprehensive dataset suitable for AI-based applications. This section outlines the proposed generative AI system designed to simulate the cardiac US environment and presents the dataset used to develop the DRL framework aimed at enabling autonomous US scanning.

\subsection{Dataset} 
\label{subsec: dataset}

Training DRL-based systems requires a simulated environment due to the large number of trial-and-error iterations involved. In the context of cardiac US, acquiring real human data is heavily constrained by ethical and regulatory limitations. To overcome these challenges, tools like the CAE Blue Phantom are used to simulate and mimic the human heart, allowing for the development, training, and evaluation of AI-based solutions.

To build a suitable simulation environment, multiple US scans of the phantom are collected, forming a dataset designed to support both generative modeling and DRL-based navigation. For each image in the dataset, a corresponding set of parameters is recorded. These include probe parameters and robotic parameters, which represent the physical configuration during image acquisition. Additionally, each image is annotated with a class label and a quality grade to support DRL objectives. The following sections describe the dataset in detail, structured around its use for generative AI and DRL-based navigation tasks.

\subsubsection{Data Acquisition For Image Generation}
\label{subsubsec: Data Acquisition For Image Generation}

\begin{figure}[H]
    \begin{center}
    \centering
\includegraphics[width=0.6\linewidth]{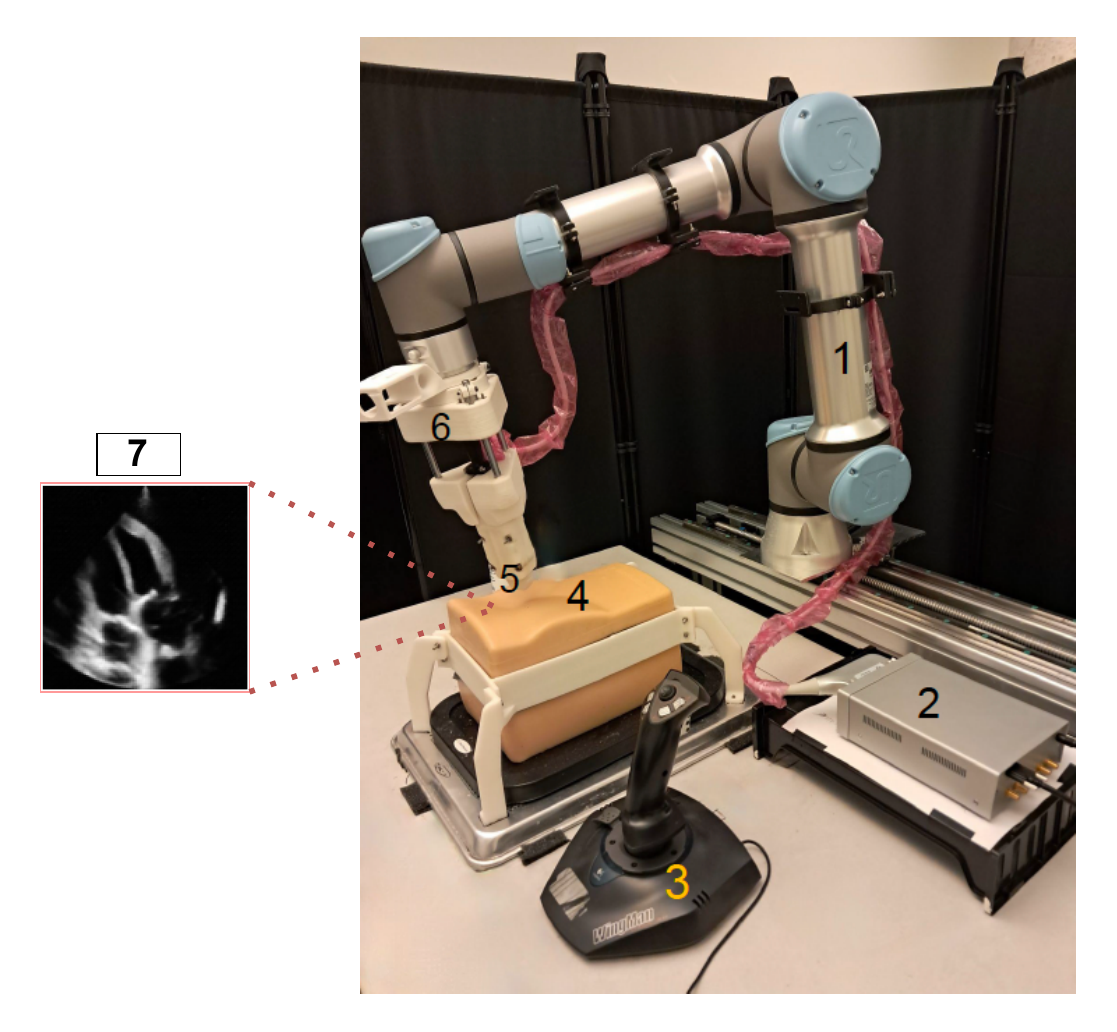}
    \caption{Illustration of the CAE Blue phantom robotic scanning setup: (1) Robotic arm, (2) Ultrasound machine, (3) Joystick, (4) Cardiac phantom, (5) Ultrasound probe, (6) End-effector, (7) Example of a scanned image}
    \label{fig:setup}
    \end{center}
    \vspace{-1em}
\end{figure}

To construct the dataset used in this work, a robotic arm was employed to perform controlled US scans on the CAE Blue Phantom. This setup enables precise control over probe movement and positioning, ensuring consistency and repeatability across scans as illustrated in Figure \ref{fig:setup}. The US scanning of the phantom was performed using the Telemed-P5-1S15-A6 in conjunction with the Telemed ArtUs-H2 US system Probe\footnote{\url{https://www.telemedultrasound.com/en/artus-scanner/}}. The UR5e robotic arm was used to manipulate the probe during data acquisition. Known for being both lightweight and highly versatile, the UR5e is widely adopted in research environments due to its ease of integration and precise motion capabilities \footnote{\url{https://www.universal-robots.com/fr/produits/ur5e/}}. Throughout the scanning process, the robotic arm was manually guided using a joystick to position the probe at various orientations and locations over the phantom surface. 

\begin{figure}[H]
    \begin{center}
    \centering
\includegraphics[width=1.0\linewidth]{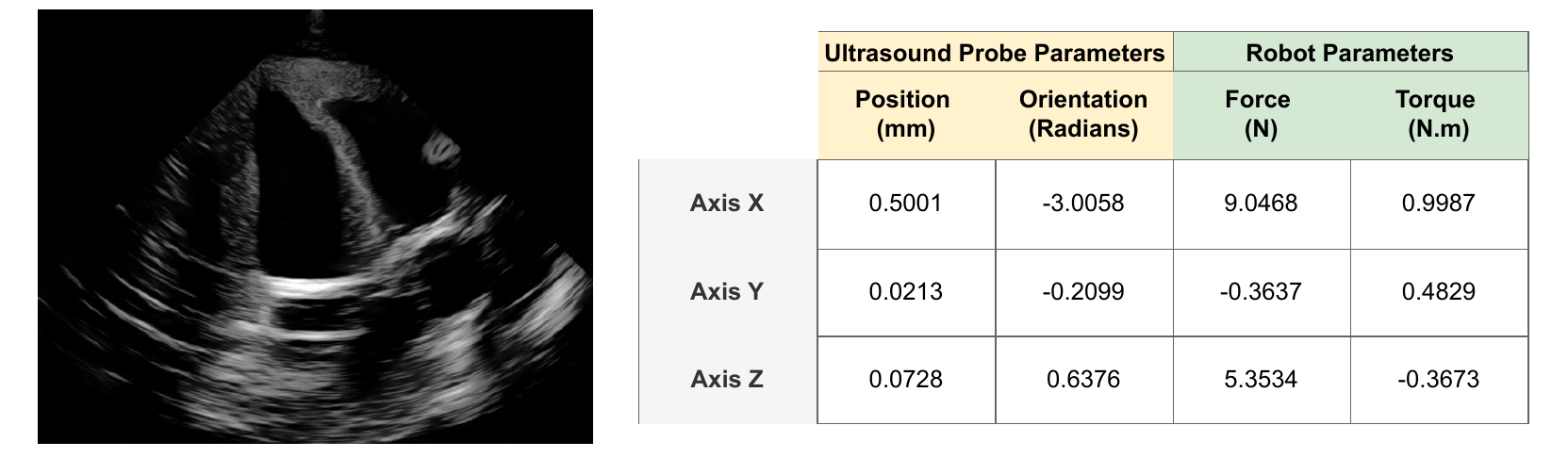}
   \caption{Illustrative example from the dataset}
    \label{fig:sample_dataset}
    \end{center}
    \vspace{-2em}
\end{figure}

For the purpose of training the DRL agent, it is essential to provide, for every point in the environment, the current state, the action taken, and the resulting next state. To support this requirement, data were acquired in real time and organized into multiple trajectories. Each trajectory comprises synchronized US images, corresponding probe positions and orientations, as well as the applied force and torque measurements. A sample image from the dataset, along with its associated parameters, is shown in Figure \ref{fig:sample_dataset}. Note that the orientation values in the table are represented in Euler angles; the units of the force parameters are Newtons, torque is measured in Newton-meters, and position is given in millimeters. The US system parameters remained fixed during the entire data acquisition process to ensure consistency.
All the data are saved to a computer linked to the US machine. In the context of this study, each trajectory is designed to capture one of five standard cardiac views: apical four chamber (A4C), subcostal four chamber (SC), parasternal long axis (PL), and two parasternal short axis views targeting the aortic valve (PSAV) and the mitral valve (PSMV). This dataset is specifically used for image generation task, where the goal is to train the generation model to synthesize US images based on input parameters such as probe pose and applied force. The goal is for the model to learn the underlying mapping between physical input parameters and the corresponding image outputs.

\subsubsection{Data Acquisition For Autonomous US Scanning}
\label{subsubsec: Data Acquisition For Autonomous Ultrasound Scanning}

To support the development of the DRL model, an image quality assessment model was also designed to assist the robotic system during the scanning process. This DL-based system aims to classify and grade US images. Notably, the same US images used for image generation are repurposed for both the classification and grading tasks. Each image is categorized into one of the standard cardiac view classes: A4C, SC, PL, PSAV and PSMV. Images that do not correspond to any of these standard views are labeled as ``Random".
In addition to classification, each image is assigned a quality score ranging from 1 (poor quality) to 10 (excellent quality), based on the completeness and clarity of the anatomical structures. Images labeled as ``Random" were assigned a grade of 0, as they do not represent a valid cardiac view. This classification and grading process is beneficial for training DL models, as it enables the system to distinguish between relevant views and irrelevant images in real time during scanning. The grading criteria used were established by cardiovascular imaging experts and are described in detail in \cite{elmekki2025cactus}.


\subsubsection{Dataset Statistics}
\label{subsubsec: Dataset_statistique}

\begin{figure}[H]
    \begin{center}
    \centering
\includegraphics[width=1.0\linewidth]{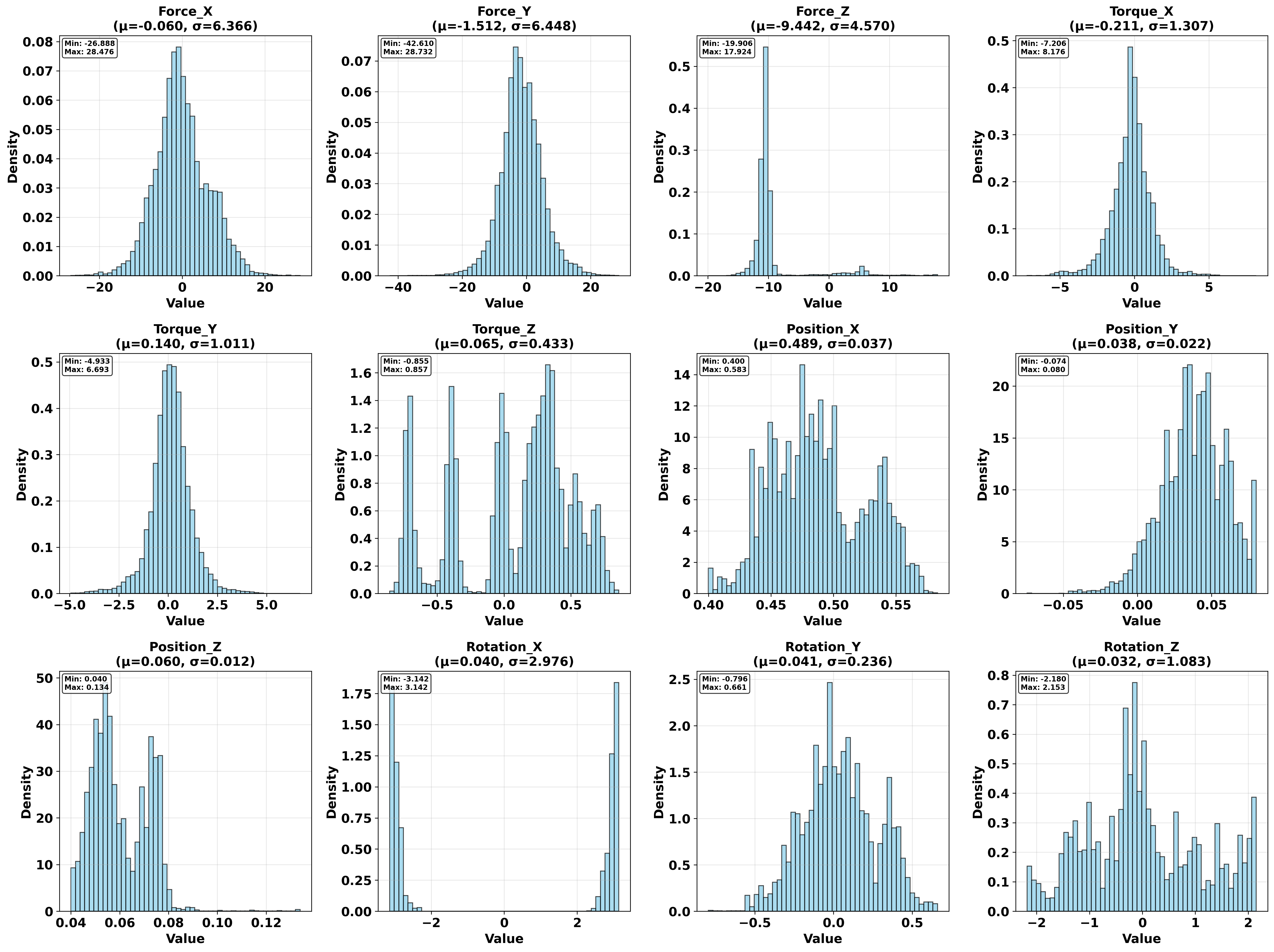}
   \caption{Parameters distribution (Units – Force: Newtons, Torque: Nm, Position: mm, Rotation: Radians)}
    \label{fig:parameters_distribution}
    \end{center}
    \vspace{-2em}
\end{figure}

The dataset used in this study comprises 144, 668 samples. Each sample contains an US image along with 12 associated parameters: \textit{Force\_X}, \textit{Force\_Y}, \textit{Force\_Z}, \textit{Torque\_X}, \textit{Torque\_Y}, \textit{Torque\_Z}, \textit{Position\_X}, \textit{Position\_Y}, \textit{Position\_Z}, \textit{Rotation\_X} (Pitch), \textit{Rotation\_Y} (Roll), and \textit{Rotation\_Z} (Yaw). The force and torque components represent the linear (\textit{Force\_X}, \textit{Force\_Y}, \textit{Force\_Z}) and rotational (\textit{Torque\_X}, \textit{Torque\_Y}, \textit{Torque\_Z}) forces exerted by the US probe on the cardiac phantom, as measured by a force sensor. The position and rotation parameters correspond to the translational and rotational vectors, respectively, indicating the displacement and orientation of the US probe relative to the cardiac phantom.
The table below \ref{tab:parameters_statistics} presents statistical information for each parameter, including the minimum, maximum, mean, and standard deviation (Std) values. Additionally, Figure \ref{fig:parameters_distribution} presents the distribution of the various parameters across the dataset. Most of the parameters follow an approximately Gaussian distribution with some fluctuations. However, the parameter \textit{Rotation\_X} (Pitch) exhibits a distinct bimodal distribution, suggesting reduced variability compared to the other rotational parameters. \textit{Torque\_Z} also shows a multimodal distribution, characterized by multiple peaks. In the case of \textit{Force\_Z}, the values are predominantly negative, which aligns with the scanning context where the movement generally involves applying downward pressure on the phantom during the US acquisition. The dataset can be accessed via this link: \href{https://1drv.ms/u/c/fb338ea7cf297329/EfVSxRaamsBJptZZQvQL_x4BriOIntoxHnmSBDwc5Hxb1Q?e=JhOklb}{\underline{RACINES Dataset}}.


\begin{table}[h]
  \centering
  \begin{tabular}{|c |c|c|c|c|}
\hline
    \textbf{Parameter} & \textbf{Min}  & \textbf{Max} & \textbf{Mean} & \textbf{Std} \\ \hline
Force\_X (N) & -26.8882 & 28.4758 & -0.0600 & 6.3660 \\
\hline

Force\_Y (N) & -42.6100 & 28.7321 & -1.5118 &  6.4483 \\
\hline

Force\_Z (N) & -19.9065 & 17.9240 & -9.4422 & 4.5702 \\
\hline

Torque\_X (N.m) & -7.2058 & 8.1759 & -0.2110 & 1.3068 \\
\hline

Torque\_Y (N.m) & -4.9334 & 6.6932 & 0.1395 & 1.0109 \\
\hline

Torque\_Z (N.m) & -0.8552 & 0.8573 & 0.0650 & 0.4330 \\
\hline

Position\_X (mm) & 0.3999 & 0.5833 & 0.4886 & 0.0375  \\
\hline

Position\_Y (mm) & -0.0744 & 0.0801 & 0.0380 & 0.0216  \\
\hline

Position\_Z (mm) & 0.0400 & 0.1339 & 0.0602 & 0.0115  \\
\hline

Rotation\_X (Radians) & -3.1416 & 3.1416 & 0.0399 & 2.9759  \\
\hline

Rotation\_Y (Radians) &  -0.7959 & 0.6613 & 0.0412 & 0.2355   \\
\hline

Rotation\_Z (Radians) & -2.1800 & 2.1534 & 0.0317 & 1.0826  \\
\hline

\end{tabular}
  \caption{Parameter statistics}
\label{tab:parameters_statistics}
\vspace{-1em}
\end{table}

\subsection{Continous Conditional US Image Generation}
\label{subsec: Continous Conditional Ultrasound Image Generation}

One of the primary objectives of this study is to develop a realistic simulation environment for cardiac US imaging using a cardiac phantom, aimed at supporting the development of DRL-based solutions for autonomous US scanning. To achieve this, generative AI techniques were explored and GANs emerging as a particularly suitable choice. GANs are widely recognized as a state-of-the-art approach in generative modeling. Their architecture, composed of a generator and a discriminator operating in an adversarial framework, fosters a competitive learning process that encourages the generator to produce increasingly realistic outputs.

GANs have demonstrated a strong ability to generate high-resolution, high-fidelity images which is an essential requirement in medical imaging. Another key advantage of GANs lies in their architectural flexibility, which allows for customization and the integration of auxiliary modules tailored to specific applications. These strengths collectively justify the adoption of a GAN-based solution in the context of cardiac US simulation.

In this study, we specifically employ a ccGAN architecture to generate US images conditioned on continuous variables. These conditioning inputs include probe position, orientation, applied force, and torque. However, standard cGANs, which typically perform well with discrete class labels, often fail to offer fine-grained control when conditioning on continuous variables. Conventional methods, such as concatenating conditioning inputs with noise vectors or appending them at the discriminator’s final layer, have shown limited effectiveness. These strategies frequently resulted in the generation of nearly identical images despite varying input conditions, due to a weak conditioning signal. Additionally, challenges such as mode collapse and the discriminator overpowering the generator during training further hindered the performance of baseline GAN architectures.

\begin{figure}[H]
    \begin{center}
    \centering
\includegraphics[width=1.0\linewidth]{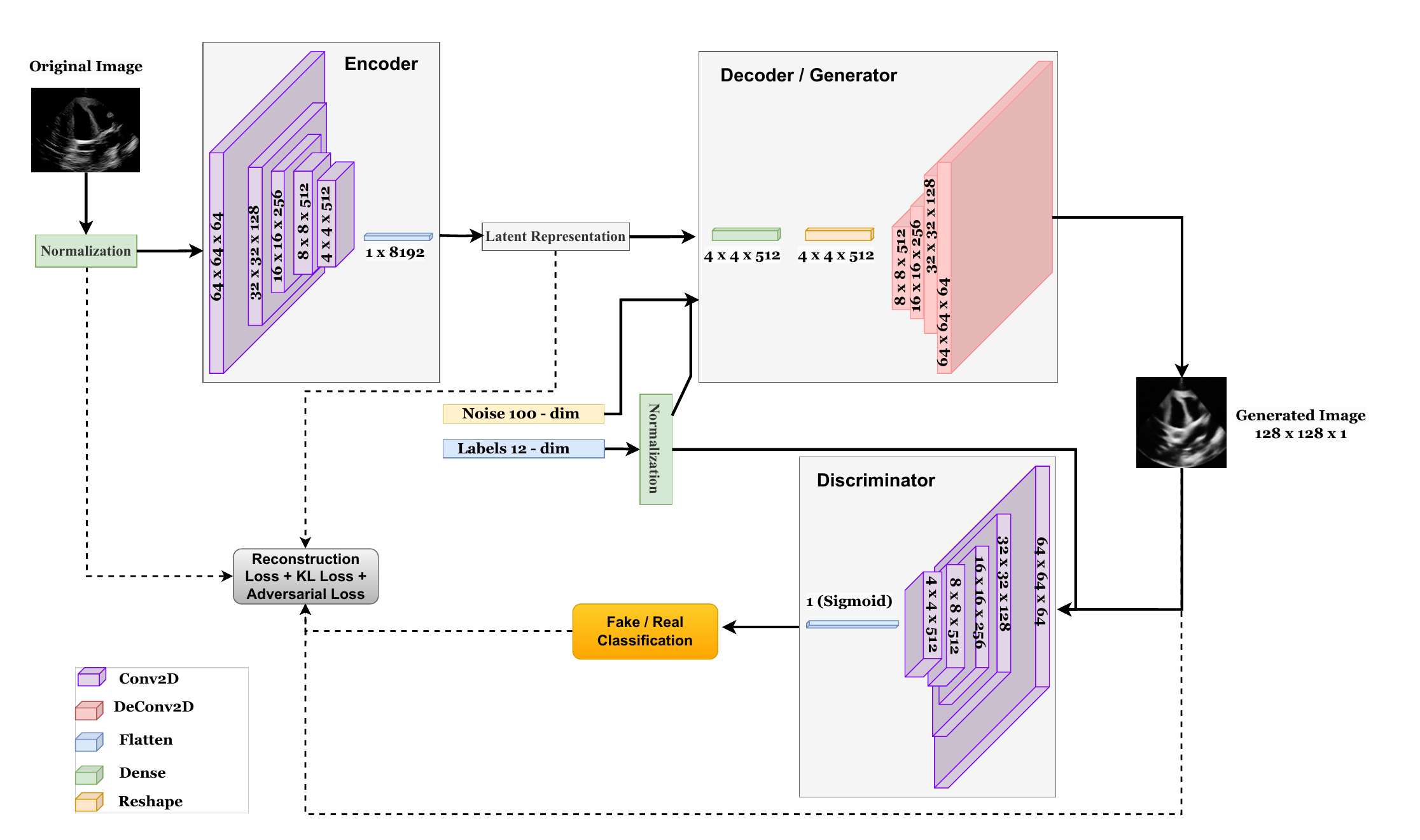}
    \caption{Architecture of the continous conditional US image generation system}
    \label{fig:vaegan}
    \end{center}
    \vspace{-2em} 
\end{figure}

To overcome these limitations, we propose a hybrid architecture that integrates a VAE with the ccGAN, as illustrated in Figure \ref{fig:vaegan}. The VAE module is used to learn the latent representation of the US data. The encoder, composed of 2D convolutional layers followed by a flattening layer, extracts latent features from the input cardiac US images. This latent representation is then passed to the ccGAN's generator, which serves a dual purpose: acting as a VAE decoder and as a GAN generator. The decoder reconstructs the image using a sequence of dense, reshape, and deconvolutional layers. Simultaneously, the generator synthesizes new images based on random noise and the continuous probe parameters. 

The objective of the VAE is to facilitate the generation of diverse images conditioned on different input parameters, thereby addressing the issue of generating identical images for varying conditions. Instead of injecting the condition into multiple layers of the GAN, the VAE preserves image diversity by incorporating the condition through concatenation with the latent vector at the input stage of the generator. In the discriminator, the image is processed independently through convolutional layers, while the labels are passed through a separate embedding network. Both outputs are concatenated only at the final layer.

The discriminator evaluates the generated images by jointly processing the images and their corresponding conditioning labels. Its architecture consists of multiple convolutional layers followed by a flattening layer. To ensure training stability, improve convergence, and enhance the visual quality of the generated images, batch normalization is applied throughout all network components.

The training process optimizes multiple loss functions: a reconstruction loss to ensure fidelity between input and reconstructed images within the VAE, a Kullback–Leibler (KL) divergence loss to regularize the latent space, and an adversarial loss to encourage the generator to produce realistic and condition-specific images. This integrated framework enables high-quality cardiac US image generation guided by continuous probe parameters, addressing the shortcomings of traditional GAN approaches in medical simulation contexts.

\section{DRL-Based Autonomous US Scanning}
\label{Sec: drl_based_autonomous_ultrasound_scanning}

The second component of the proposed solution is a DRL-based system designed for autonomous US scanning. DRL is particularly well-suited for tasks that involve continuous decision-making and interaction with complex environments such as navigating the probe over a dynamic anatomical structure like the heart. Through trial-and-error interactions and feedback, DRL enables agents to learn optimal probe trajectories, resulting in reliable and precise decision-making strategies. Given the requirement for real-time responsiveness and high precision in US scanning, DRL offers an AI-based framework for learning safe, effective, and autonomous navigation behaviors. This section provides a detailed description of the DRL-based solution and its integration within the overall system.

\subsection{Policy Architecture}
\label{subsec: Policy Architecture}

In DRL, the environment is represented based on the Markov decision process (MDP) defined by :
$\mathcal{S}$ the set of states, $\mathcal{A}$ the action space, p with $p\left(s_{t+1} \mid s_{t}, a_{t}\right)$ is the stochastic dynamics between states for a given action, $\mathcal{R}(s, a)=r \in \mathbb{R}$ the reward function and $\gamma \in[0,1)$ the discount factor. In reinforcement learning, an agent is trained following the policy $\pi(a \mid s)$ which can maximize the expected reward $r$ by selecting an action $a_{t}$, based on the policy parameterized by $\theta$ and defined as $\pi_{\theta}(a \mid s)$.

\begin{figure}[H]
    \begin{center}
    \centering
\includegraphics[width=1.0\linewidth]
{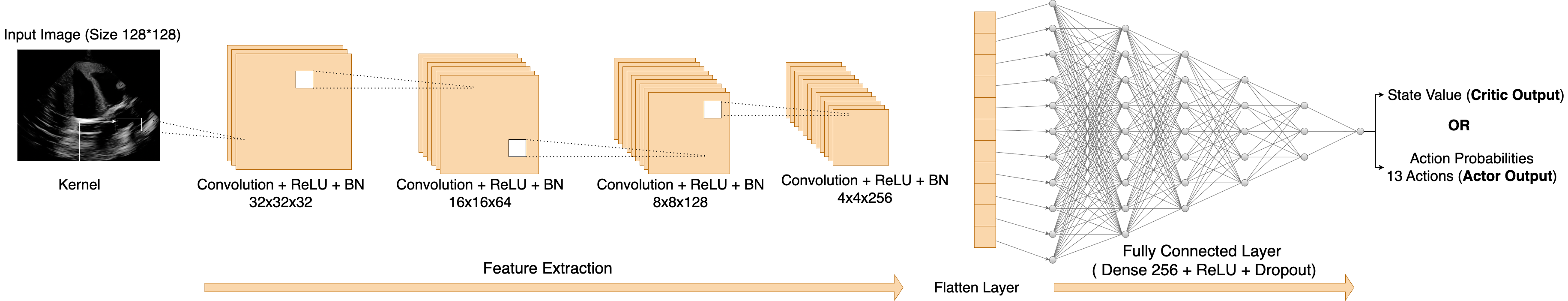}
   \caption{Architecture of the actor and critic networks}
    \label{fig:actor_critic}
    \end{center}
\end{figure}

In the proposed work, PPO is adopted due to its demonstrated performance across a variety of environments, including multi-agent systems, robotics, vehicle control, and gaming \cite{elumalai2025proximal,alagha2025uav,maldonado2021visual, alagha2025adaptive, sami2023reward}. PPO is particularly known for its training stability, which stems from its clipped surrogate objective and its actor-critic framework. The algorithm relies on two neural networks: an actor, which proposes actions based on the current state, and a critic, which estimates the value of those states. Figure \ref{fig:actor_critic} illustrates the common architecture of both the actor and critic networks implemented in this study. Notably, the two networks share an identical structure, beginning with a sequence of convolutional layers combined with batch normalization to extract relevant features from the input US images. These features are then passed through fully connected layers. The actor network outputs a probability distribution over possible actions, while the critic outputs a scalar value representing the expected return of the current state.

\subsection{State \& Action Spaces}
\label{subsec: State and action spaces}

In the context of this study, the state is represented by the cardiac US image generated by the generative system described in Section \ref{subsec: Continous Conditional Ultrasound Image Generation}. At each time step \(t\), the agent observes an image \(I_{t}\), which encodes the current state of the environment as a vector of raw pixel intensities. The action space \(\mathcal{A}\) consists of 13 actions, including 12 discrete actions defined as a combination of translations and rotations along and about the three spatial axes:

\begin{equation*}
\mathcal{A} = \{ \pm T_x, \pm T_y, \pm T_z, \pm R_x, \pm R_y, \pm R_z , Idle \} \tag{1}
\end{equation*}

\noindent where \(T_x, T_y, T_z\) represent unit translations along the x, y, and z axes, respectively, and \(R_x, R_y, R_z\) denote unit rotations about those same axes. Each action moves the probe in the physical space, guiding the agent toward acquiring the target cardiac view. The 13th action is an idle action, in which the agent remains stationary. This action is used when the agent encounters boundary regions, helping to train the agent to avoid violating spatial constraints.

\subsection{Reward Function}
\label{subsec: Reward function}

Several DRL-based US navigation methods use sparse or distance-based reward functions, where rewards are calculated either based on the distance between the agent’s current and target positions or on image similarity measures. However, such methods lack generalizability, as the exact spatial coordinates of the target view are typically unknown in real clinical scenarios. In contrast, in the context of cardiac US, the target view is known and sonographers are trained to identify and acquire specific anatomical views based on image content, not fixed spatial positions. Therefore, the robot should be trained to think in terms of visual goals, mimicking the sonographer’s reasoning.

To address this, the reward function used in this study is inspired by the grading and classification framework introduced in \cite{elmekki2025cactus}. This framework employs a CNN based on ResNet-18, which was initially trained for view classification and later fine-tuned using transfer learning to assess image quality. The primary aim of employing transfer learning is to reduce computation time, the number of training parameters, and floating point operations by using a common encoder with separate heads, rather than using two separate models. The resulting quality scores, which reflect the diagnostic value of the images, are used as feedback signals for the DRL agent. These scores enable the agent to learn navigation strategies that prioritize reaching views with high clinical relevance, rather than simply minimizing spatial distance. The grading system was established by medical imaging experts and is based on two criteria: completeness and clarity of the cardiac US image. The proposed reward function is defined as follows:

\[
R_t = R_{\text{base}} + R_{\text{class}} + R_{\text{grade}} + R_{\text{step}}
\tag{2}
\]

\noindent where \( R_{\text{base}} \) is the reward that compares the current classification probability \( p_t \) of the current state with the threshold 0.9, and also compares the current grade \( g_t \) of the current state with the threshold 5.0, which represents the midpoint of the grading scale defined in \cite{elmekki2025cactus}, ranging from 0 to 10. Both the classification probability and grade are outputs of the classification-grading framework developed:

\[
R_{\text{base}} = \begin{cases} 
20.0 & \text{if } p_t \geq 0.9 \text{ and } g_t < 5.0 \\
50.0 & \text{if } p_t \geq 0.9 \text{ and } g_t \geq 5.0 \\
0 & \text{otherwise}
\end{cases} \tag{3}
\]

The objective of this reward is to assess whether the agent successfully reaches the anatomically target cardiac view with high confidence. A high reward of 50 is assigned when the agent not only reaches the target view but also achieves high image quality according to the predefined scoring schema, which evaluates the view based on completeness and clarity. If the agent reaches the correct cardiac window with high confidence but the image quality is suboptimal and the assigned grade is below 5.0, the agent receives a reward of 20. This lower reward is intended to encourage the agent to reach the correct view while still motivating improvement in image quality. The values 20 and 50 were selected through empirical tuning. The goal of these relatively high reward values is to strongly incentivize the agent for reaching the target view, while maintaining a moderate ratio between high- and low-quality outcomes (i.e., 50 vs. 20). This setup also helps keep cumulative episode returns within a controlled range, which provides a clear learning signal and contributes to value function stability

\( R_{\text{class}} \) compares the current classification probability \( p_t \) with the previous step's probability \( p_{t-1} \):

\[
R_{\text{class}} = p_t - p_{t-1} \tag{4}
\]

This reward \( R_{\text{class}} \) encourages the agent, at each step, to seek improved cardiac views based on the implemented classification mechanism. This reward shaping approach provides continuous feedback throughout the agent's trajectory, rather than only at the target view.

We also apply a reward shaping mechanism for the grading component, where  \( R_{\text{grade}} \) provides intermediate rewards at each step during the learning episode. \( R_{\text{grade}} \) compares the current grade \( g_t \) with the previous grade \( g_{t-1} \), but only when the classification confidence is high; if the classifier is not confident in the current class, the grade difference is not considered:

\[
R_{\text{grade}} = \begin{cases} 
g_t - g_{t-1} & \text{if } p_t \geq 0.9 \\
0 & \text{otherwise}
\end{cases} \tag{5}
\]

Finally, \( R_{\text{step}} \) is a small negative constant used to penalize long episodes, encouraging the DRL agent to reach the target in the minimum number of steps possible.

The full training algorithm and implementation details of the DRL framework, including reward computation and policy optimization, are presented in Algorithm \ref{alg:ppo_training}. This algorithm describes the PPO training procedure. During an episode of fixed length, at each timestep \( h \), an action \( a_h \) is selected based on the previous state \( s_{h-1} \) according to the current policy. This action is then executed in the environment, resulting in a new state \( s_h \) and a reward \( r_h \). The resulting transition is stored in a buffer \( \mathcal{B} \), and the cumulative reward for the episode is updated accordingly.
If the current timestep coincides with a scheduled policy update, the advantage estimates are computed using the collected rewards and the value function. Subsequently, both the actor and critic networks are updated. This process is repeated until all episodes have been completed.

\begin{algorithm}[htp]
\caption{Monitored PPO Training}
\label{alg:ppo_training}
\begin{algorithmic}[1]
\STATE \textbf{Input:} Environment $\mathcal{E}$, PPO agent $\pi_\theta$, hyperparameters
\STATE \textbf{Output:} Trained policy $\pi_{\theta^*}$
\STATE
\STATE Initialize $T_{\text{max}}$ \COMMENT{Maximum training timesteps}
\STATE Initialize $T_{\text{update}}$ \COMMENT{Update frequency}
\STATE Initialize $H$ \COMMENT{Maximum episode length}
\STATE $t$ \COMMENT{Current timestep}
\STATE $\mathcal{B} \leftarrow \emptyset$ \COMMENT{Experience buffer}
\STATE
\WHILE{$t \leq T_{\text{max}}$}
    \STATE $s_0 \leftarrow \mathcal{E}.\text{reset()}$ \COMMENT{Initialize episode}
    \STATE $R_{\text{ep}} \leftarrow 0$, $\text{success} \leftarrow \text{False}$
    \FOR{$h = 1$ to $H$}
        \STATE $a_h \leftarrow \pi_\theta(s_{h-1})$ \COMMENT{Select action}
        \STATE $s_h, r_h, \text{done}, \text{info} \leftarrow \mathcal{E}.\text{step}(a_h)$
        \STATE $\mathcal{B}.\text{store}(s_{h-1}, a_h, r_h, \text{done})$ \COMMENT{Store experience}
        \STATE $R_{\text{ep}} \leftarrow R_{\text{ep}} + r_h$
        \STATE $t \leftarrow t + 1$
        \IF{$t \bmod T_{\text{update}} = 0$}
            \STATE Compute the advantage estimate $\hat{A}$ using the rewards and the value function $V$ (critic)
            \STATE Update the actor (policy) and critic networks using the loss $\mathcal{L}^{CLIP+VF+S}(\theta)$
            \STATE $\mathcal{B} \leftarrow \emptyset$ \COMMENT{Clear buffer}
        \ENDIF
        \IF{\text{done}}
            \STATE $\text{success} \leftarrow \text{True}$
            \STATE \textbf{break}
        \ENDIF
    \ENDFOR
    \STATE \textbf{LogEpisode}($R_{\text{ep}}, h, \text{success}$)
\ENDWHILE
\STATE
\STATE \textbf{return} $\pi_{\theta^*}$
\end{algorithmic}
\end{algorithm}

\section{Experiments}
\label{Sec: Experiments}
This section evaluates the proposed AI system which is an end-to-end framework combining GAN and DRL for simulating and performing autonomous US scanning and designed to be adaptable and generalizable to multiple organs beyond the heart. We begin by describing the experimental setup, followed by a detailed analysis of the results, including the performance of the image generation models, the DRL-based scanning solution, and a benchmarking of different state representations. The section concludes with a qualitative assessment by medical imaging experts.
The objective is to show the effectiveness of the proposed framework and evaluate its performance through both qualitative and quantitative analyses.

\subsection{Experiments Setup}
\label{subsec: Experiments Setup}

The experiments involved multiple steps across both hardware and software components.
On the hardware side, the robotic setup is integrated with a controller, and the US probe is mounted on the robot via the end-effector. A force sensor is positioned at the end-effector to measure the applied force during operation. All communication, controller development, and parameter tuning are detailed in \cite{zakeri2024ai}. 
Using this robotic setup, we collect the data described in Section \ref{subsec: dataset}, which is used to train the proposed solution.

On the software side, several preprocessing steps were applied prior to training to ensure data consistency and improve model performance. For building the simulation environment through data generation, we have followed a data preparation and preprocessing procedure that includes resizing to 128×128 pixels, grayscale conversion, and normalization to the range [–1, 1] using a mean and standard deviation of 0.5. All 12 conditioning values including probe position, orientation, and applied forces were normalized to create a uniform dataset. Orientation data were initially collected as pose matrices and subsequently converted to Euler angles prior to normalization. For the image quality assessment model related to building the reward function, the same preprocessing steps were applied.

For training VAE and GAN, the hyperparameter configuration employed a batch size of 8, an image size of 128 × 128 pixels, a latent vector dimension of 100. Training was performed for 100 epochs with a learning rate of 0.0001.

For the DRL training, the environment was set with a maximum episode length of 200 steps, and policy updates were performed every 2,048 timesteps. Validation was carried out every 10,000 timesteps over 100 episodes. The PPO algorithm is updated after every 5 training epochs, a clipping parameter of 0.2, an actor learning rate of 0.00002, and a critic learning rate of 0.0001. Training continued for a total of 15 million timesteps, with a discount factor of 0.95. All experiments were conducted on the Compute Canada Narval cluster \footnote{\url{https://docs.alliancecan.ca/wiki/Narval/en}}.

\subsection{Results and Discussion}
\label{subsec: Results and Discussion}

This section presents the experiments conducted to evaluate the performance of the proposed solution. First, we assess the effectiveness of the image generation system based on the VAE-GAN architecture. This is followed by a comparative benchmark of different GAN variants, evaluated both qualitatively through visual inspection of the generated images and quantitatively using standard image quality metrics, including Fréchet Inception Distance (FID) \cite{heusel2017gans}, Structural Similarity Index Measure (SSIM) \cite{wang2004image}, and Peak Signal-to-Noise Ratio (PSNR). SSIM reflects the brightness, contrast, and structural similarity between the synthesized US image and the reference image—the higher the SSIM value, the better the synthesized image quality. FID measures the distance between the real and synthesized images at the feature level, providing an indication of perceptual similarity.
Subsequently, we evaluate the performance of the DRL-based autonomous US scanning system, comparing different state representations and reward function designs. Finally, an expert assessment is provided by medical imaging professionals to validate the quality of the generated and acquired US views.

\subsubsection{Performance of the Image Generation Models}
\label{subsubsec: Performance of GenAI}

\begin{figure}[htp]
    \begin{center}
    \centering
\includegraphics[width=0.7\linewidth]{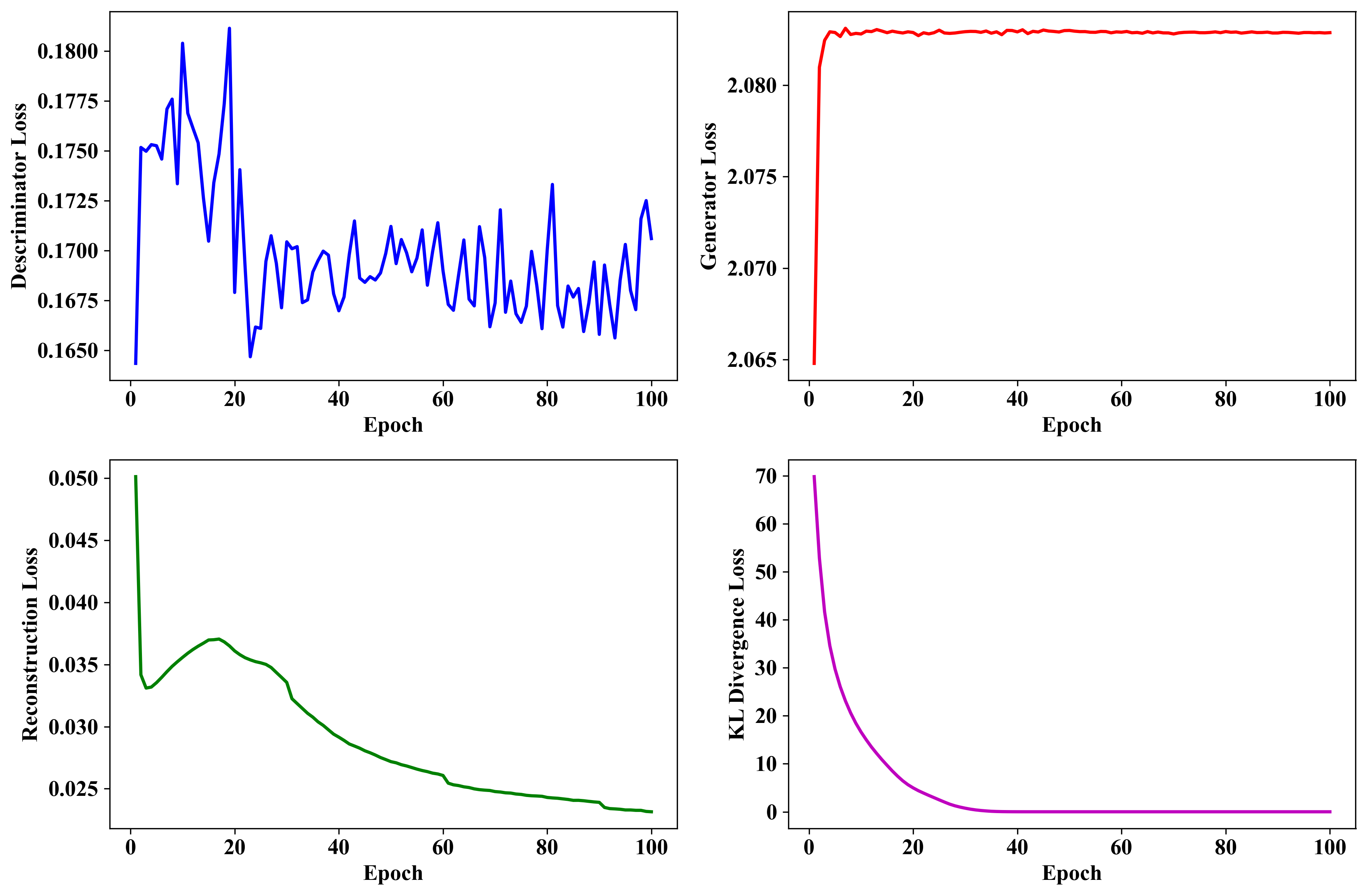}
    \caption{Training losses of the VAE-GAN model}
    \label{fig:vaegan_losses}
    \end{center}
    \vspace{-2em}
\end{figure}

The VAE-GAN model demonstrates high performance in generating high-quality US images, as shown in Figure \ref{fig:vaegan_losses}. During training, the discriminator loss converged around epoch 100, reaching a value of 0.3251, while the generator loss also stabilized at approximately the same point with a value of 2.0822. The discriminator loss, being neither close to zero nor excessively high, demonstrates  that the discriminator is not overly confident in distinguishing real from generated images, nor is it completely failing. The convergence of the generator loss implies that the generator has learned to produce realistic US images capable of challenging the discriminator effectively, without being overpowered by it. Similarly, both the reconstruction loss and the KL divergence loss exhibited steady convergence, reaching minimal values of 0.0230 and 0.0013, respectively, around epoch 100. These results show that the autoencoder is capable of reconstructing images that closely resemble the original inputs, while also maintaining a well-regularized latent space. Overall, these outcomes indicate the high performance of the VAE-GAN model and its effective learning of both the data distribution and the underlying latent space representation.

To further assess the performance of our proposed model, we conducted a comparative benchmark with several GAN-based architectures: baseline cGAN, conditional StyleGAN inspired from \cite{karras2019style}, attention cGAN, and residual cGAN. Each variant was conditioned on continuous inputs, which were integrated into all layers of the architecture to ensure consistent control over the generation process. The objective of this benchmark was to evaluate how architectural enhancements such as attention \cite{vaswani2017attention} and residual connections \cite{he2016deep} affect image quality in the context of conditioned cardiac US generation.

\begin{figure}[h]
    \begin{center}
    \centering
\includegraphics[width=0.6\linewidth]{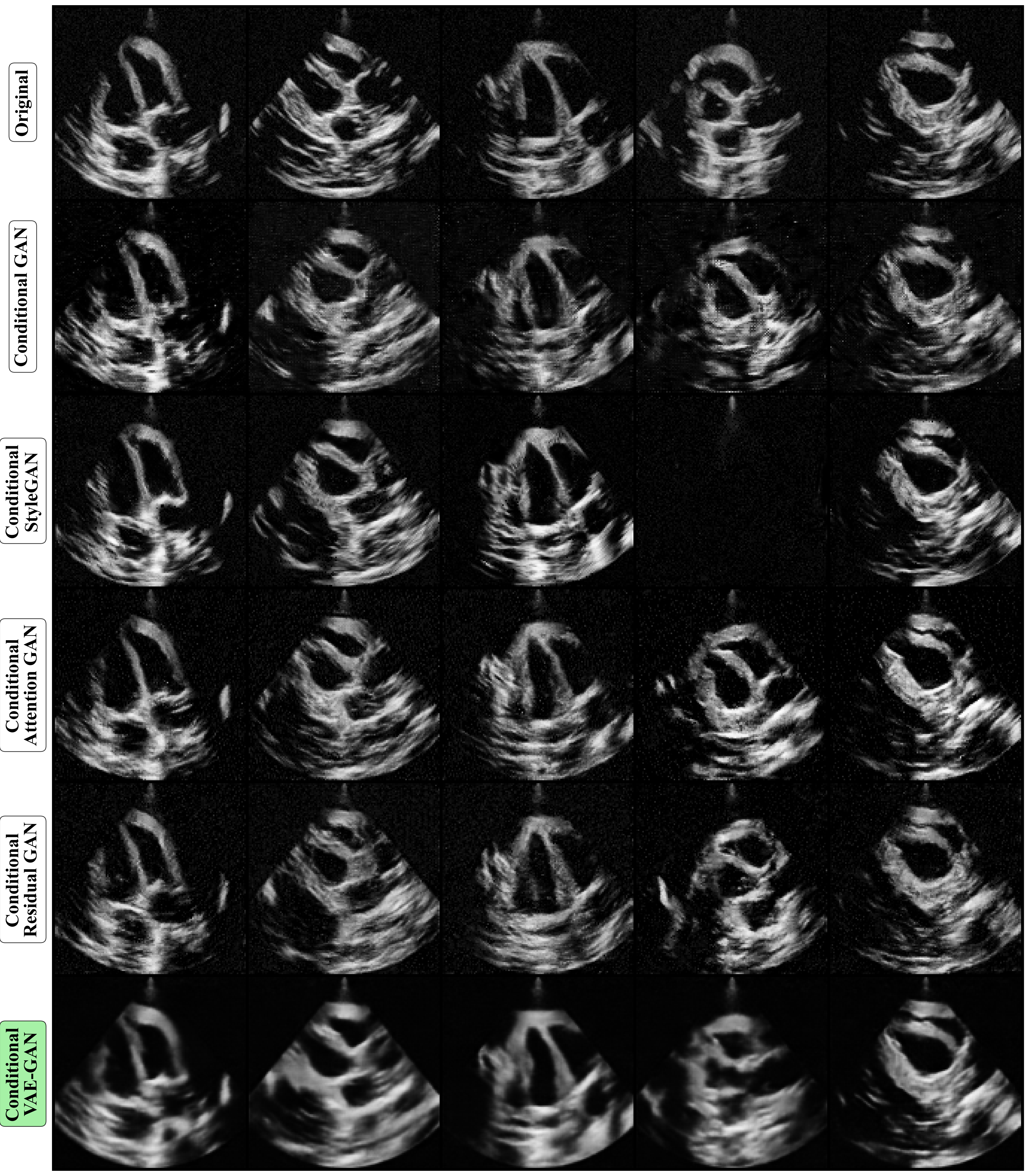}
    \caption{Qualitative evaluation of conditional GAN benchmarks. \textit{Note: The black image shown for Conditional StyleGAN is the output produced by the model itself, not a visualization error}}
    \label{fig:genai_comparison}
    \end{center}
    \vspace{-2em} 
\end{figure}

A Qualitative comparison, illustrated in Figure \ref{fig:genai_comparison} shows that the conditional VAE-GAN produces images that are more similar to the original US scans. For instance, in some cases, the StyleGAN failed entirely to generate certain images, producing black outputs (e.g., in the fourth sample). Other variants generated images with a rough anatomical structure but suffered from noticeable noise and missing details. Although the conditional VAE-GAN also exhibited limitations, particularly in the fourth example, where it failed to reproduce certain anatomical structures and it generally produced clearer and more accurate images compared to other approaches.

\begin{figure}[h]
    \begin{center}
    \centering
\includegraphics[width=1.0\linewidth]{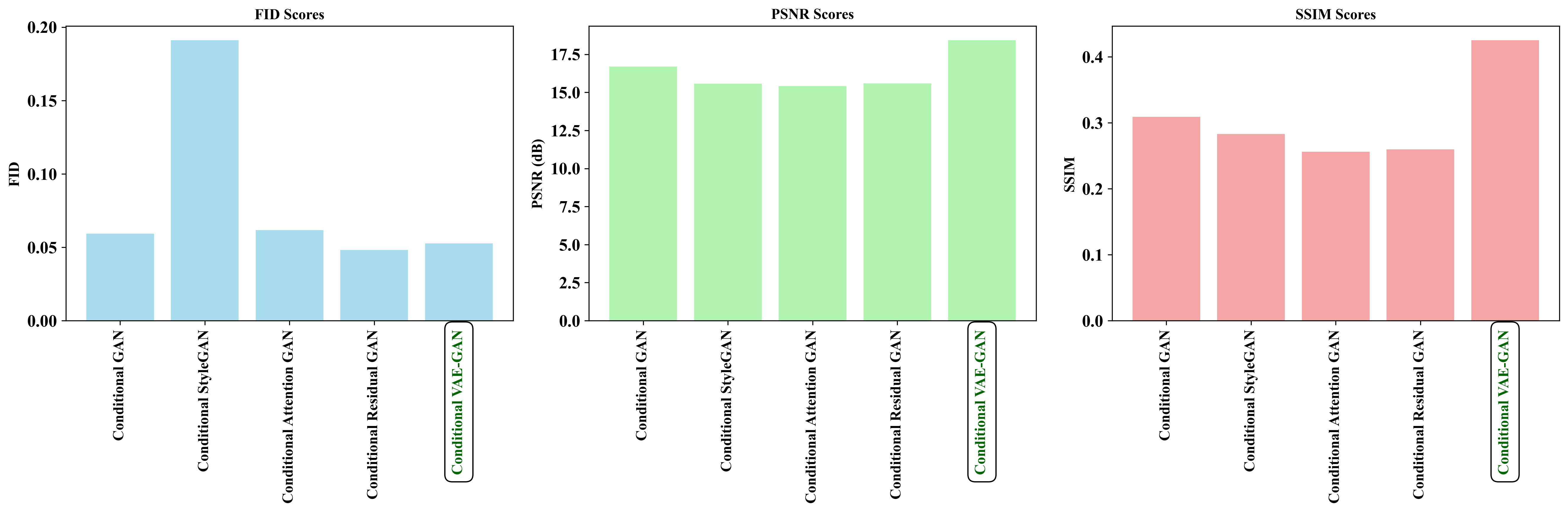}
    \caption{Quantitative evaluation of conditional GAN benchmarks}
    \label{fig:genai_metrics}
    \end{center}
    \vspace{-1em} 
\end{figure}

Quantitative results, presented in Figure \ref{fig:genai_metrics}, further supports these findings. As shown in the figure, the proposed conditional VAE-GAN demonstrates superior performance compared to the other models in terms of SSIM, achieving a score of 0.4253, and PSNR, with a value of 18.4312. A higher SSIM and lower PSNR indicate that the generated images are structurally more similar to real images and exhibit less noise. However, in terms of FID, the conditional VAE-GAN ranked third with a score of 0.0526, following the attention-based cGAN (0.0594) and the residual cGAN (0.0483). These results suggest that while the VAE-GAN offers strong structural fidelity and noise suppression, other architectures may better capture high-level semantic features when evaluated using feature distribution metrics such as FID.

\subsubsection{Performance of the Autonomous US Scanning}
\label{subsubsec: performance_autonomous_us_scanning}

In this context, the scanning process begins from a randomized starting position in each episode to increase the robustness and generalizability of the proposed DRL-based solution. Since the heart has multiple standard views, the results presented focus on the convergence behavior toward a specific target view—namely, the SC cardiac view.

\begin{figure}[h]
  \begin{subfigure}{.30\textwidth}
  \centering
    \includegraphics[width=1\linewidth]{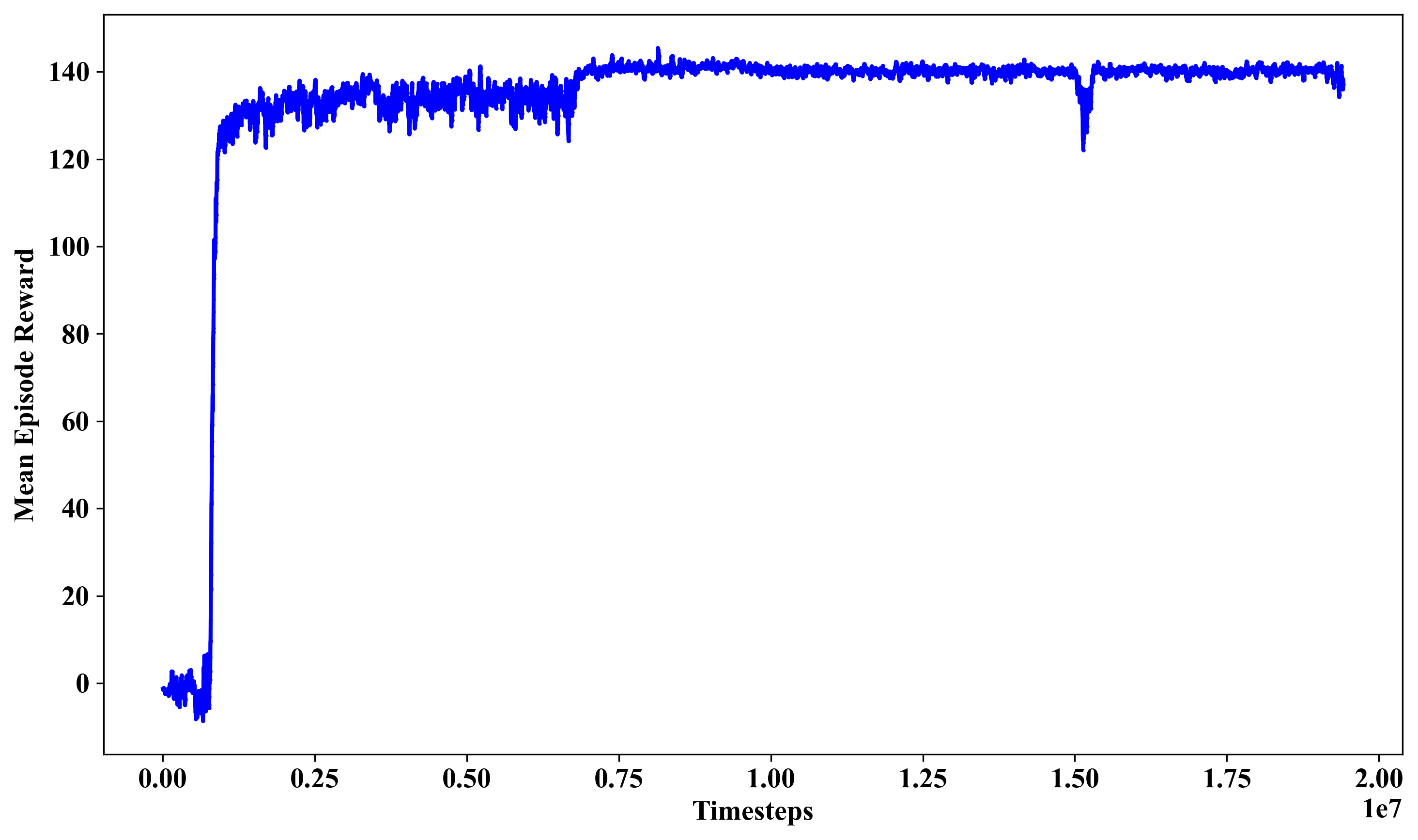}
    \caption{Training mean \\ reward}
  \end{subfigure}%
  \hspace{1em} %
  \begin{subfigure}{.30\textwidth}
  \centering
    \includegraphics[width=1\linewidth]{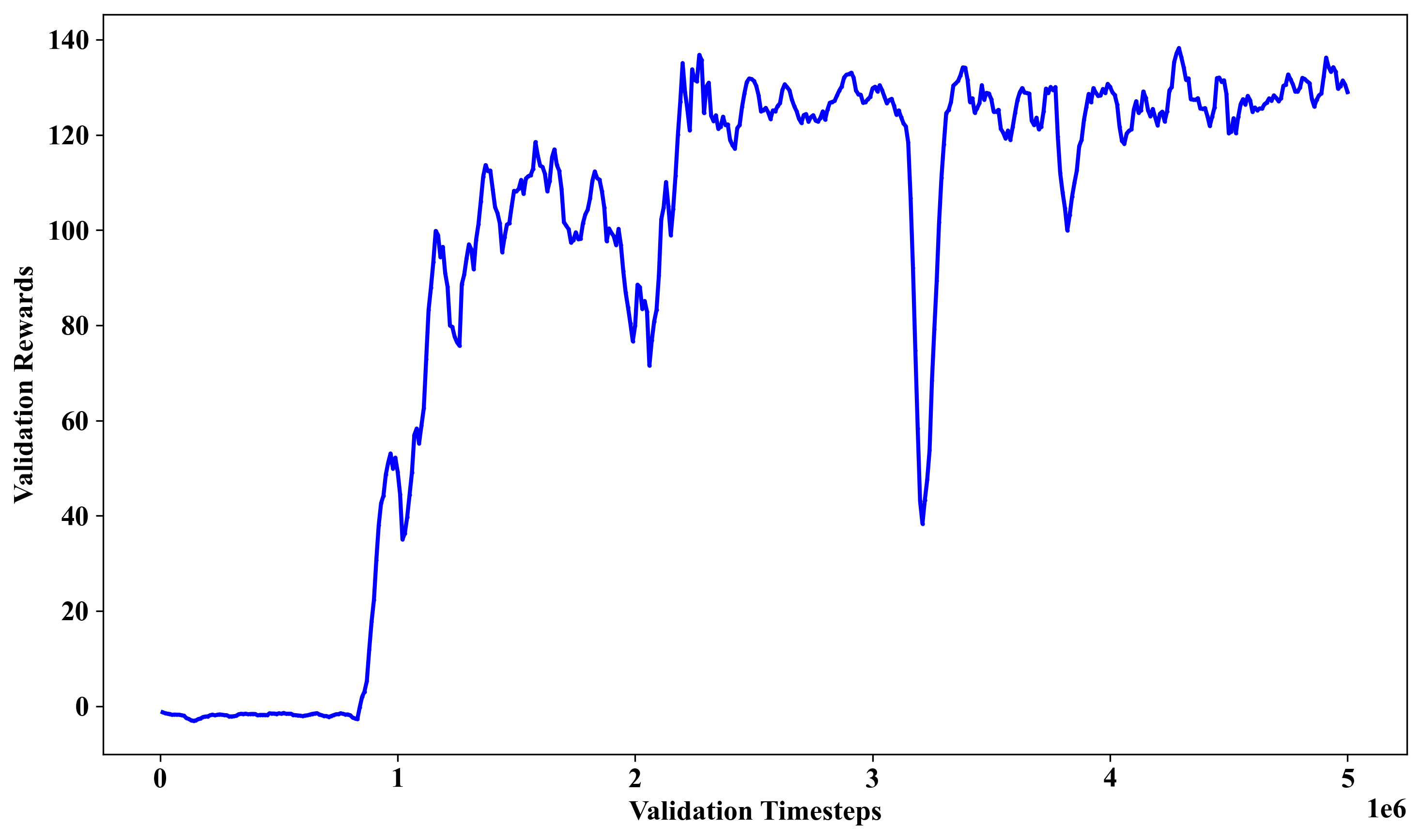}
    \caption{Validation mean reward}
  \end{subfigure}%
  \hspace{1em} %
  \begin{subfigure}{.30\textwidth}
  \centering
    \includegraphics[width=1\linewidth]{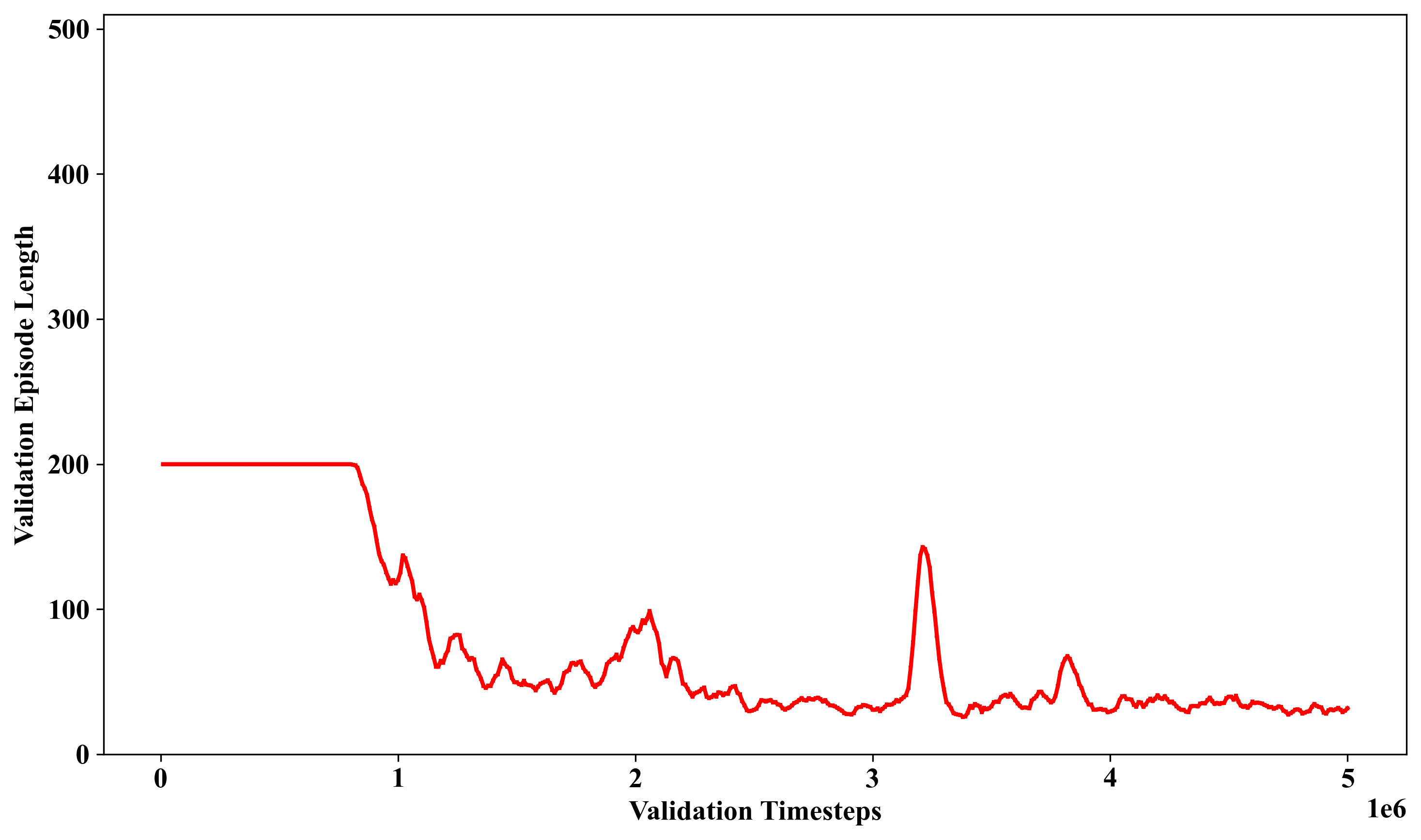}
    \caption{Validation episode length}
  \end{subfigure}
  \caption{Training \& validation performance}
  \label{fig:only_image}
  \vspace{-1em}
\end{figure}

The evaluation of the DRL-based navigation system was carried out through several performance metrics. As shown in Figure \ref{fig:only_image} (a), the mean training reward begins to converge around timestep 7,500,000, stabilizing at an average reward of approximately 140. Additionally, validation performance, evaluated every 10,000 timesteps across 100 episodes, confirms convergence trends in both reward (Figure \ref{fig:only_image} (b)) and episode length (Figure \ref{fig:only_image} (c)). The episode length consistently decreases over time, indicating more efficient navigation by the agent. However, occasional peaks in episode length were observed. These can be attributed to the randomness of the initial probe position, as certain starting points require more steps to reach the target cardiac view.

\begin{figure}[htp]
    \begin{center}
    \centering
\includegraphics[width=0.6\linewidth]{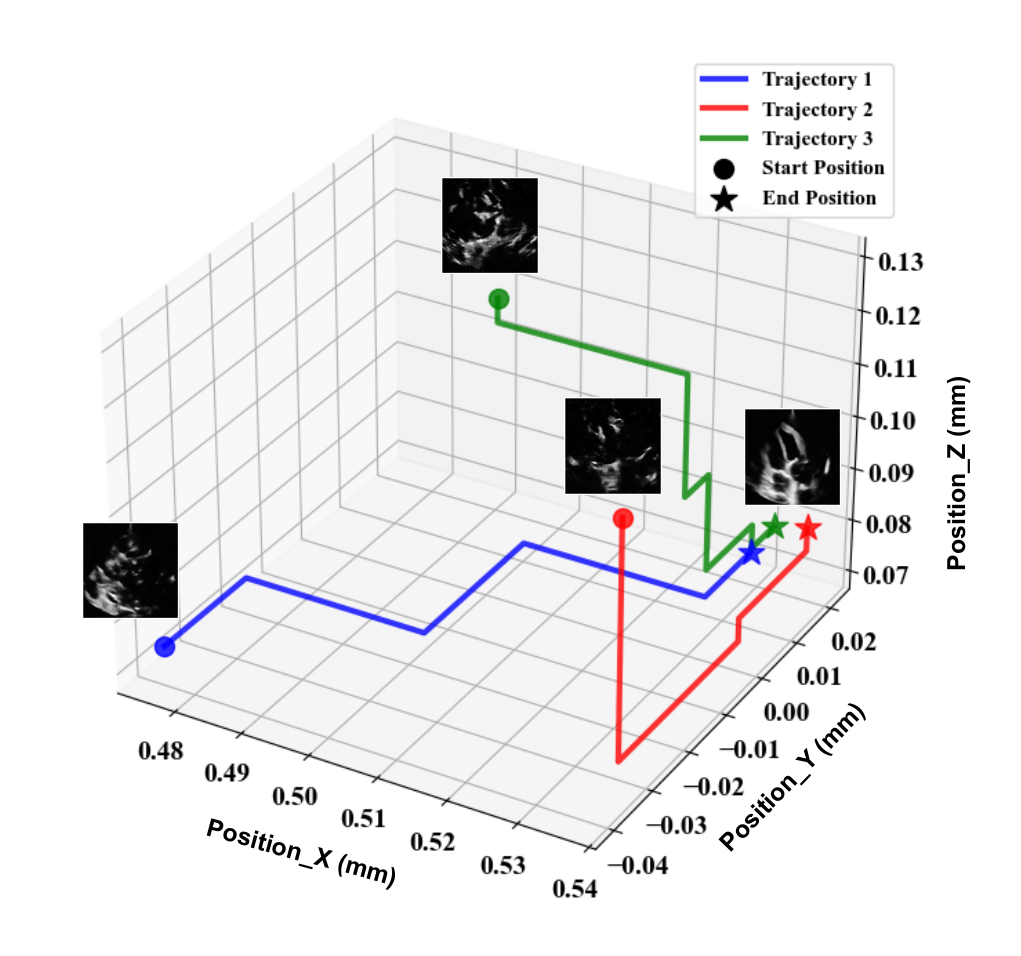}
    \caption{Trajectory visualization from multiple random initial points during inference}
\label{fig:trajectory}
    \end{center}
    \vspace{-2em} 
\end{figure}

To further demonstrate the ability of the agent to efficiently reach the target view from various starting positions and within a limited number of steps, we recorded its behavior (i.e. positions and actions taken). During inference, as illustrated in Figure \ref{fig:trajectory}, the trained agent consistently reaches the SC view from various random starting points. In the blue trajectory, the agent achieves the target view in 16 steps with a classification confidence of 0.9547, completing the trajectory in 0.36 seconds. In the red trajectory, it reaches the same anatomical view from a different starting point in 32 steps, with a confidence of 0.9265 and a total time of 0.56 seconds. In the third trajectory, the agent requires 22 steps, completing the task in 0.42 seconds with a confidence score of 0.9882. These results demonstrate both the efficiency and accuracy of the agent in consistently achieving the desired anatomical view, regardless of the initial position.

\begin{figure}[H]
    \begin{center}
    \centering
\includegraphics[width=0.8\linewidth]{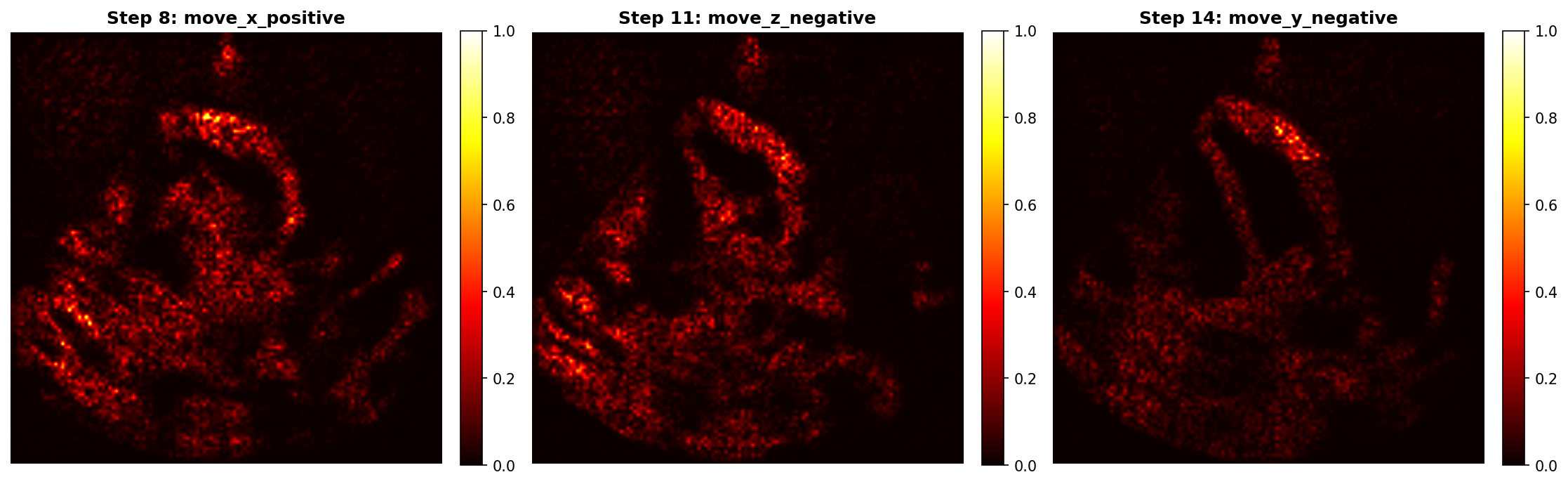}
    \caption{Analysis of DRL model attention through heatmaps across consecutive states}
\label{fig:last_three_heatmaps}
    \end{center}
    \vspace{-2em}
\end{figure}

Given the promising results described above, it is important to understand the internal decision-making processes of the model. This is particularly important considering the long-term goal of deploying the system in clinical settings, including trials involving human subjects. In this context, explainable AI plays a key role, as it provides tools and methods to interpret and analyze how models behave in response to input data. From this perspective, Integrated Gradients, an explainable AI technique \cite{sundararajan2017axiomatic}, is used to understand the relationship between a DL model’s input and its corresponding predictions. As shown in Figure \ref{fig:last_three_heatmaps}, this method highlights the regions of the US image that are most influential in guiding the decision-making process of the DRL model. The figure presents three consecutive states sampled at intervals of three steps. Each image corresponds to a specific time step within an episode, along with the associated probe movement. In step 8, the movement is a translation along the positive x-axis. The second image corresponds to step 11, where the movement is along the negative z-axis. Finally, the third image corresponds to step 14 and results from a translation along the negative y-axis. To generate the heatmaps for each step, the following procedure was applied. For each US image at a given time step, 50 interpolated images were generated, starting from a fully black baseline image and gradually blending in the actual image. For instance, the first interpolation consists of 2\% of the original image and 98\% black, while the final interpolation step represents 100\% of the original image. A gradient is computed for each interpolated image, and the average gradient is then calculated. This average is multiplied by the difference between the actual image and the black baseline to produce the final attribution heatmap. 

It can be observed that the agent consistently focuses on the anatomical structures within the US images. These areas, represented by brighter regions, have a stronger impact on the model’s action selection, whereas darker areas indicate minimal influence. This suggests that the DRL agent, trained using only US images as input states, is capable of effectively guiding the scanning process based on relevant anatomical features.

\subsubsection{Performance of Different DRL State Representations}
\label{subsubsec: Computation analysis}
This section analyzes the effect of different state representations on the performance of the proposed DRL-based US scanning system. In existing studies on autonomous cardiac US scanning, particularly those focused on TTE, state representations vary, including the use of the position of the US probe \cite{shida2024robotic}, 2D US images extracted from 3D volumes \cite{shen2023towards}, and the relative positions of anatomical reference points between the current and target views \cite{lin2023deep}. The objective of this section is to benchmark various state representation configurations and evaluate their impact on system performance to determine the most effective approach. Three state configurations were evaluated: an image-only state, a parameter-only state, and a multimodal state combining both. In the image-only setting, the agent receives the US image as its sole observation, as described in the previous section. The parameter-only state, on the other hand, consists of six numerical values representing the probe’s position and orientation in 3D space. Finally, the multimodal state fuses both visual data and numerical parameters to provide a richer input representation. To accommodate these different state inputs, the architecture of the actor and critic networks was adjusted accordingly. For the parameter-only configuration, a multilayer perceptron (MLP) was employed. This network processes the six input parameters through several dense layers and outputs either action probabilities (actor) or a scalar value estimate (critic). For the multimodal configuration, a more complex network was designed. Here, a CNN processes the US image into a compact feature representation, while a parallel MLP processes the six numerical parameters. These two outputs are then concatenated to form a combined feature vector, which is used to predict the agent’s actions. Despite the increased complexity and feature space size in the multimodal model, it did not surpass the performance of the simpler image-only model, highlighting the effectiveness of visual inputs for this task.

Aside from input representation and architectural design, all other components of the DRL framework, such as the training algorithm, reward function, and action space, remain consistent across all experiments. This controlled setup ensures that the observed differences in performance are attributable solely to the choice of state representation.

\begin{figure}[H]
  \begin{subfigure}{.30\textwidth}
  \centering
    \includegraphics[width=1.0\linewidth]{image_training_episode_rewards_timesteps.png}
    \caption{Image-only state}
  \end{subfigure}%
  \hspace{1em} %
  \begin{subfigure}{.30\textwidth}
  \centering
    \includegraphics[width=1\linewidth]{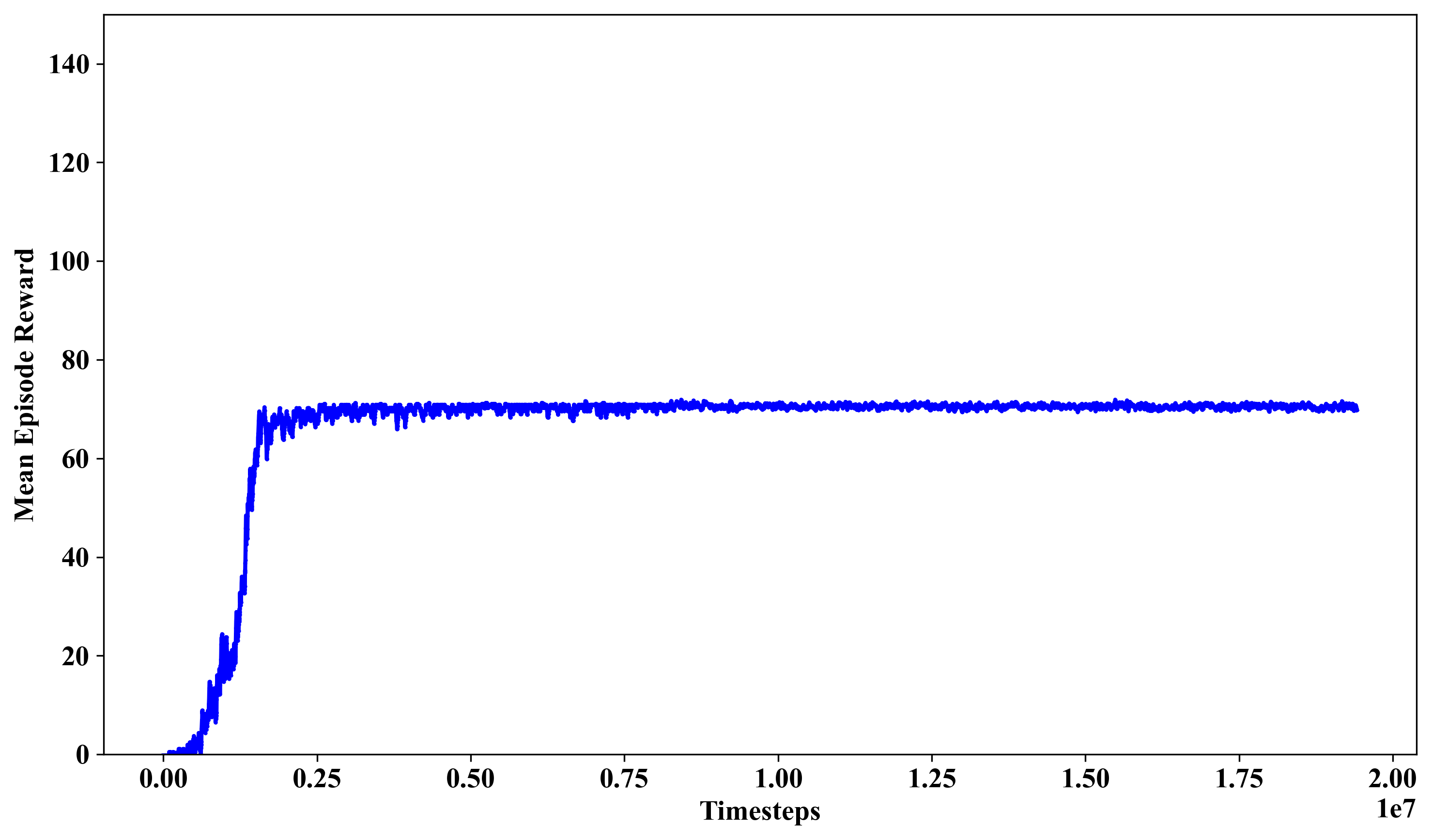}
    \caption{Parameter-only state}
  \end{subfigure}%
  \hspace{1em} %
  \begin{subfigure}{.30\textwidth}
  \centering
    \includegraphics[width=1\linewidth]{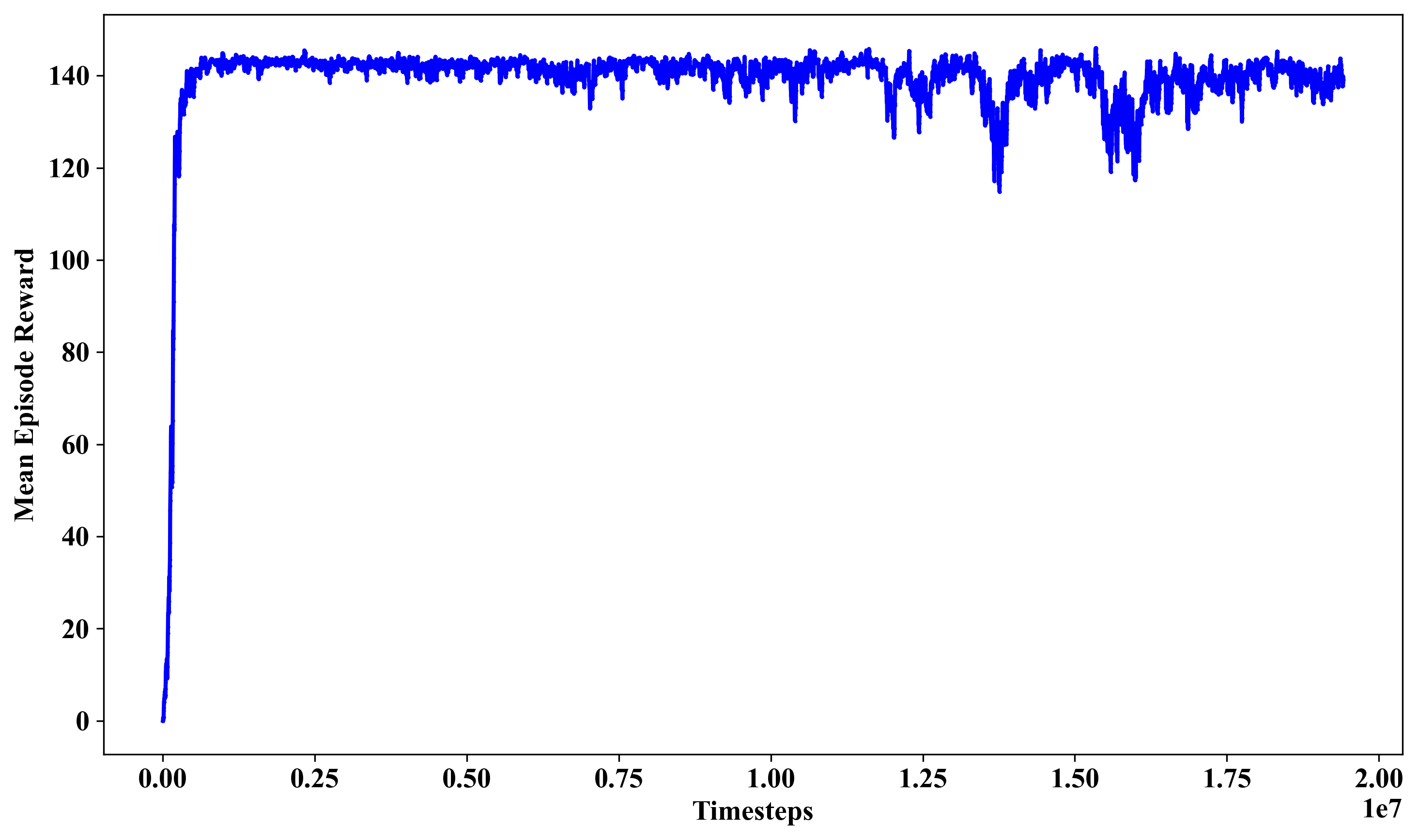}
    \caption{Multimodal-only state}
  \end{subfigure}
  \caption{Comparison of mean training rewards}
  \label{fig:training_reward_comp}
\end{figure}

\begin{figure}[H]
  \begin{subfigure}{.30\textwidth}
  \centering
    \includegraphics[width=1\linewidth]{image_validation_rewards.png}
    \caption{Image-only state}
  \end{subfigure}%
  \hspace{1em} %
  \begin{subfigure}{.30\textwidth}
  \centering
    \includegraphics[width=1\linewidth]{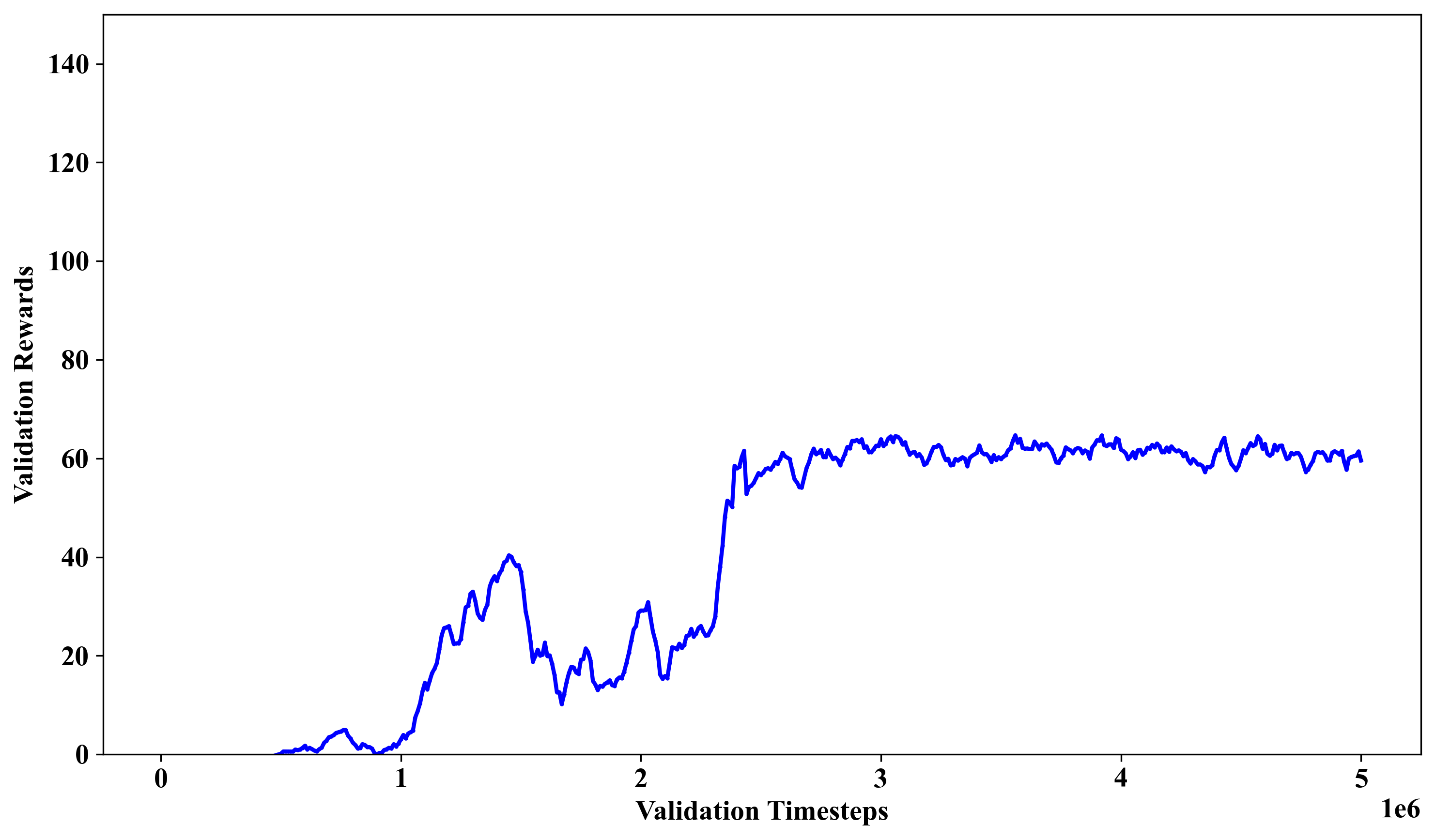}
    \caption{Parameter-only state}
  \end{subfigure}%
  \hspace{1em} %
  \begin{subfigure}{.30\textwidth}
  \centering
    \includegraphics[width=1\linewidth]{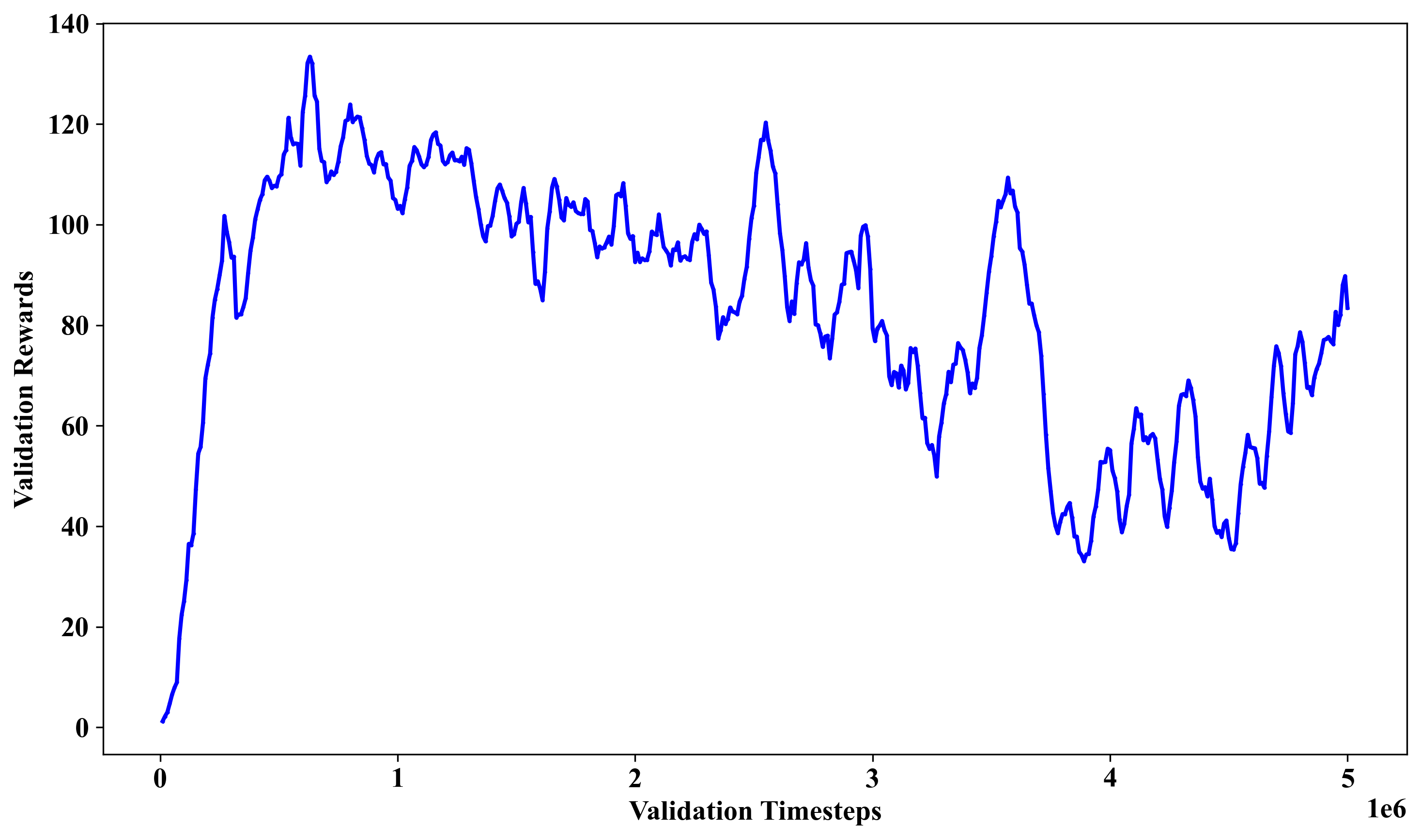}
    \caption{Multimodal-only state}
  \end{subfigure}
  \caption{Comparison of mean validation rewards}
  \label{fig:validation_reward_comp}
\end{figure}

\begin{figure}[H]
  \begin{subfigure}{.30\textwidth}
  \centering
    \includegraphics[width=1\linewidth]{image_validation_episode_lengths.png}
    \caption{Image-only state}
  \end{subfigure}%
  \hspace{1em} %
  \begin{subfigure}{.30\textwidth}
  \centering
    \includegraphics[width=1\linewidth]{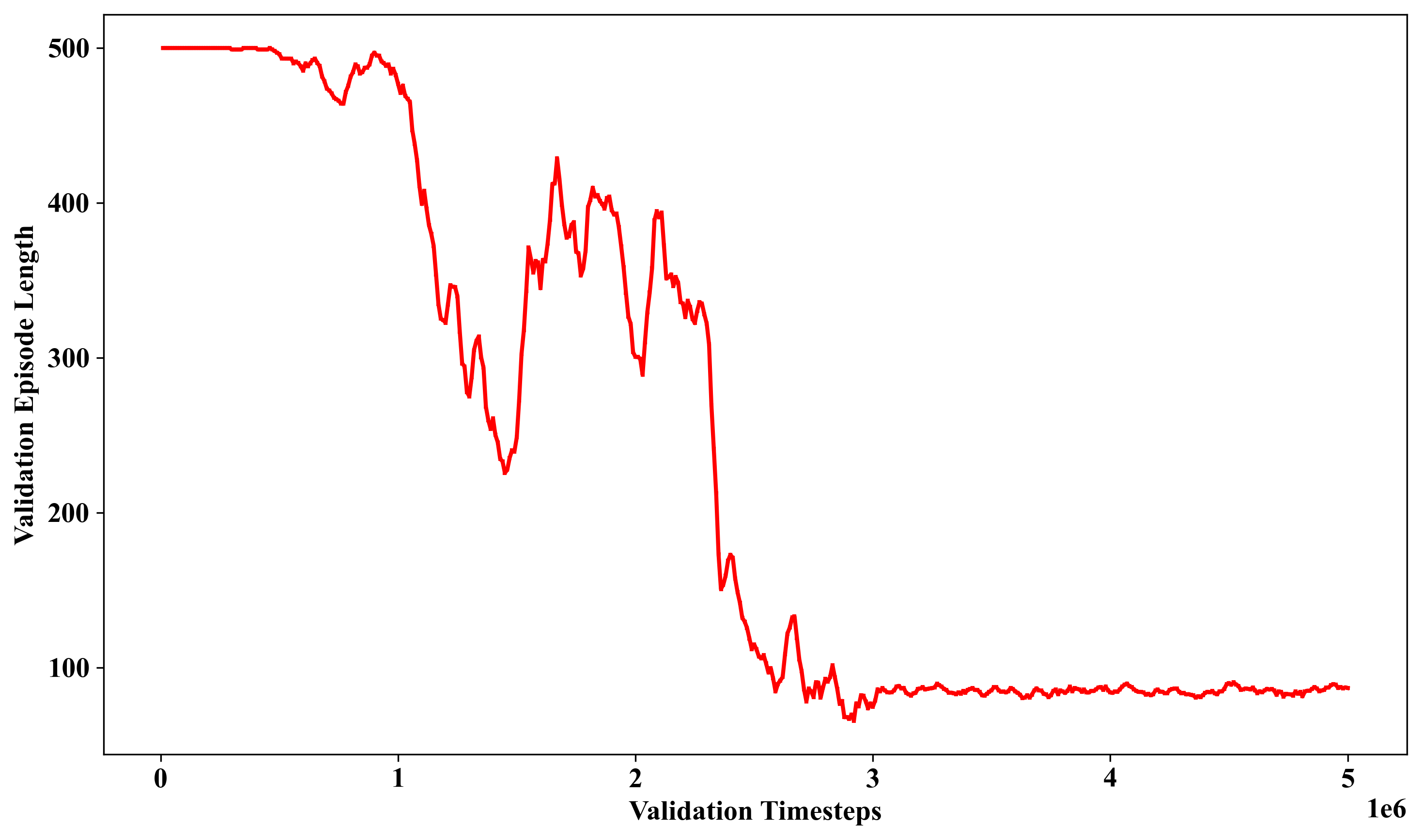}
    \caption{Parameter-only state}
  \end{subfigure}%
  \hspace{1em} %
  \begin{subfigure}{.30\textwidth}
  \centering
    \includegraphics[width=1\linewidth]{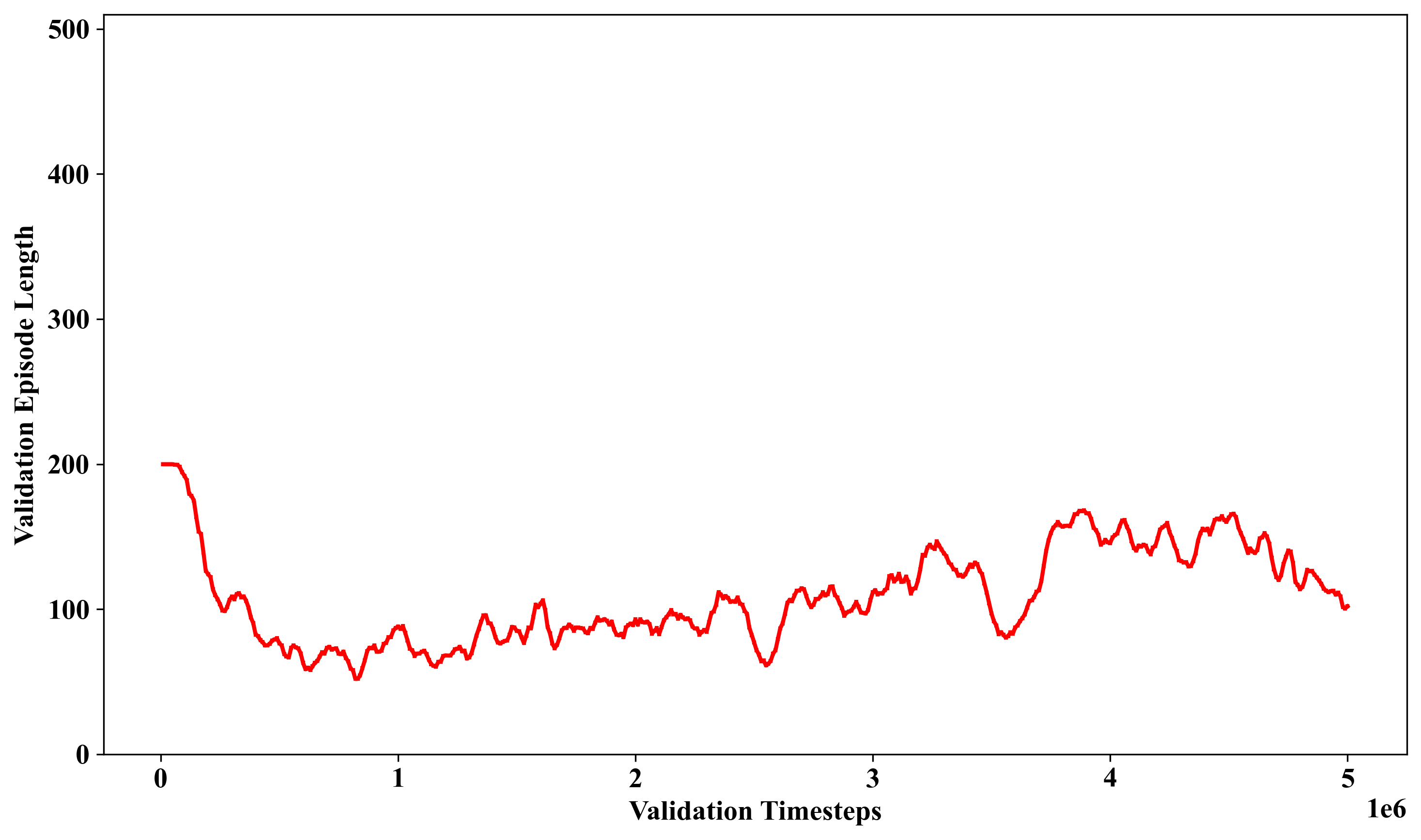}
    \caption{Multimodal-only state}
  \end{subfigure}
  \caption{Comparison of mean validation episode length}
  \label{fig:episode_length_comp}
\end{figure}

Experimental results, illustrated in Figure \ref{fig:training_reward_comp}, demonstrate that the image-only state yields superior performance. It achieves a higher mean training reward of approximately 140, compared to 70 for the parameter-only configuration. Moreover, it converges more rapidly and displays less variability than the multimodal configuration, which, while achieving a slightly similar mean training reward than the image-only state, exhibits more oscillations during training. During validation as shown in Figures \ref{fig:validation_reward_comp} and \ref{fig:episode_length_comp}, this configuration also performs best, with shorter episode lengths and higher reward values, indicating that the agent is able to reach the target cardiac view more efficiently. Specifically, in Figure \ref{fig:validation_reward_comp}, the mean validation reward reaches 140, similar to the mean training reward, suggesting consistent high performance across both training and validation. In contrast, the parameter-only configuration struggles, attaining a mean validation reward of approximately 60. The multimodal configuration performs slightly better but still deteriorates to a mean reward of around 100. A similar performance is observed in episode lengths: the image-only configuration achieves the shortest average episode length (around 25), whereas both the parameter-only and multimodal configurations result in longer episodes, with minimum lengths around 100. These findings suggest that visual feedback alone provides more informative cues for navigation than probe parameters or a combined input. In other words, the agent learns more effectively from image-based states, aligning well with how human sonographers primarily rely on visual inspection to guide scanning.

\subsubsection{Medical Imaging Expert Visual Assessment}
\label{subsubsec: Medical Imaging Expert Visual Assessment}

To further assess the quality of the US images generated by the VAE-GAN model, an expert evaluation was conducted by a medical imaging specialist. The expert was presented with 400 images synthesized by the generative model and was asked to perform both classification and grading tasks.

The results indicate that the expert’s classification aligned with the model’s predictions for 369 out of 400 images, and the grading matched the automated grading model for 350 out of 400 images. The observed discrepancies can be partly attributed to limitations inherent in the cardiac phantom particularly the absence of heart valves. This anatomical limitation may cause certain views, such as PL, A4C, and SC, to appear visually similar, leading to occasional misclassifications or grading inconsistencies. In future work, the use of more anatomically realistic cardiac phantoms, including those with dynamic pumping mechanisms and detailed valve structures, could improve the visual fidelity of the images and reduce such ambiguities.

\section{Conclusion}
\label{Sec: concolusion}

The objective of this study is to present an end-to-end framework that integrates generative AI and DRL for autonomous cardiac US scanning. This work addresses the current gap in reproducible and realistic frameworks for developing DRL-based solutions for autonomous cardiac US scanning. To this end, we propose a novel approach that uses a ccGAN combined with a VAE to generate realistic US images conditioned on spatial and robotic parameters. The proposed framework includes a comprehensive benchmarking of multiple generative models, evaluating their ability to produce high-quality and diverse US images. Among the tested models, VAE-GAN demonstrated the best performance, achieving a SSIM of 0.4375 and a PSNR of 18.5361.

Another key contribution of this work is the development of a DRL-based control policy for autonomous cardiac US scanning. The proposed DRL model relies solely on visual feedback as the state representation and employs an enhanced action space enabling robotic movements in 6 DOF. Additionally, the reward function incorporates both reward shaping and medical image quality assessment to guide the US scanning. The DRL agent was evaluated using randomized initial probe positions and was consistently able to reach the target view within a minimal number of steps, regardless of the starting point. We further compared the performance of models using purely visual states to those using numerical parameters and combined states (visual and numerical). Results show that visual feedback alone is sufficient to guide the DRL agent for effective cardiac scanning.

This framework has potential to be generalized to other organs. In the simulation component, organ-specific datasets and parameter tuning can be applied. For the DRL component, the state space can remain based on the current US image, with the reward shaped by corresponding image quality assessment. In future work, we plan to expand and refine the shared dataset RACINES by incorporating feedback from additional medical imaging experts. 
Furthermore, we aim to enhance the realism of the simulation environment by developing advanced cardiac phantoms that include dynamic heart motion and anatomically accurate valves.

\bibliographystyle{model1-num-names}
\bibliography{references.bib}

\end{document}